\documentclass[final,1p,times]{elsarticle}

\usepackage{amssymb}
\usepackage{amsmath}
\usepackage{amsthm}
\usepackage{xcolor}
\usepackage{rotating}
\usepackage[table]{xcolor}
\usepackage{graphicx}
\usepackage{dsfont}

\begin{document}
\begin{frontmatter}

\title{Intraday spatiotemporal PV power prediction at national scale using satellite-based solar forecast models}

\author[label1]{Luca Lanzilao}
\author[label1,label2]{Angela Meyer}
\affiliation[label1]{
    organization={School of Engineering and Computer Science, Bern University of Applied Sciences},
    addressline={Quellgasse 21},
    city={Biel},
    postcode={2501},
    state={Bern},
    country={Switzerland}
}
\affiliation[label2]{
    organization={Department of Geosciences and Remote Sensing, TU Delft},
    addressline={Stevinweg 1},
    city={Delft},
    postcode={2628 CN},
    state={South-Holland},
    country={The Netherlands}
}

\begin{abstract}
We present a novel framework for spatiotemporal photovoltaic (PV) power forecasting and use it to evaluate the reliability, sharpness, and overall performance of seven intraday PV power nowcasting models. The model suite includes satellite-based deep learning and optical-flow approaches and physics-based numerical weather prediction models, covering both deterministic and probabilistic formulations. Forecasts are first validated against satellite-derived surface solar irradiance (SSI). Irradiance fields are then converted into PV power using station-specific machine learning models, enabling comparison with production data from 6434 PV stations across Switzerland. To our knowledge, this is the first study to investigate spatiotemporal PV forecasting at a national scale. We additionally provide the first visualizations of how mesoscale cloud systems shape national PV production on hourly and sub-hourly timescales. Our results show that satellite-based approaches outperform the Integrated Forecast System (IFS-ENS), particularly at short lead times. Among them, SolarSTEPS and SHADECast deliver the most accurate SSI and PV power predictions, with SHADECast providing the most reliable ensemble spread. The deterministic model IrradianceNet achieves the lowest root mean square error, while probabilistic forecasts of SolarSTEPS and SHADECast provide better-calibrated uncertainty. Forecast skill generally decreases with elevation. At a national scale, satellite-based models forecast the daily total PV generation with relative errors below 10\% for 82\% of the days in 2019–2020, demonstrating robustness and their potential for operational use.
\end{abstract}

\begin{keyword}
Solar-energy nowcasting \sep Spatiotemporal forecasting \sep Probabilistic forecasts \sep Numerical weather prediction models
\end{keyword}
\end{frontmatter}

\begin{figure}[h]
	\centering
	\includegraphics[width=0.843\textwidth]{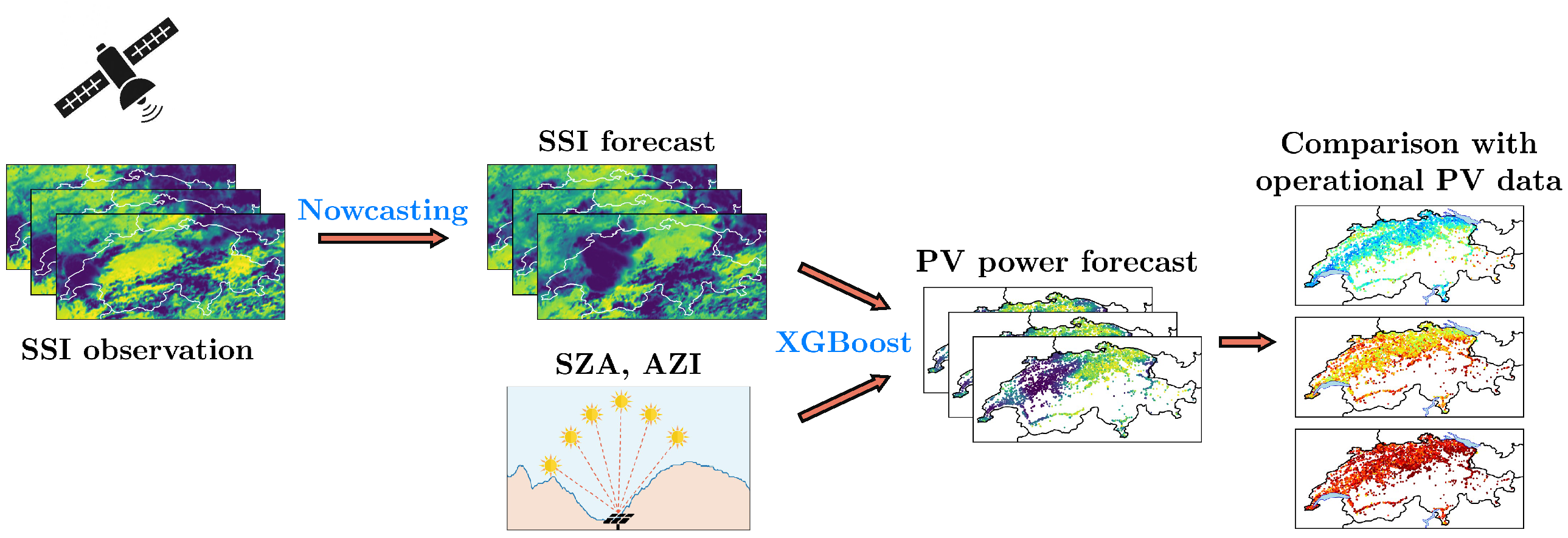}
\end{figure}

\section{Introduction}\label{sec:intro}
Solar energy plays a crucial role in decarbonising energy production and mitigating climate change. As its share in the power grid continues to grow \cite{IEA2024}, accurate forecasts of surface solar irradiance (SSI) and photovoltaic (PV) power generation are essential for grid operators, producers, and traders to ensure stability and balance supply and demand. However, intraday PV forecast models still face significant limitations. Most existing approaches still rely on deterministic methods \cite{Sharma2023, Aldahidi2025}, often based only on PV production time series \cite{Vandermeer2018, Xiang2024, Diazbello2025}. By contrast, the use of satellite data and computer vision for large-scale PV forecasting remains limited \cite{Ahmed2020, Benavides2022, Barhmi2024, Sultana2025}. Previous studies have explored forecasts combining weather and satellite data with ground station measurements \cite{Qin2022}, but these typically cover only small sets of PV sites, are deterministic, and often suffer from spatial smoothing. To properly assess the performance and robustness of spatiotemporal forecasting methods, evaluation across large PV fleets with different climate regimes and surface albedos is beneficial. However, operational datasets used so far rarely exceed a few dozen PV systems \cite{Lonij2012, Agoua2019, Karimi2021, Brester2023, Perera2024, Wang2025}. In this work, we address these gaps by introducing a novel framework for spatiotemporal PV power forecasting. We benchmark the performance of several satellite- and physics-based models and validate them against a large-scale national PV fleet comprising over 6400 PV systems. To our knowledge, this is the first study to compare satellite-based and numerical weather prediction (NWP)-based PV power forecasts using PV generation data at a national scale.

There are three main approaches to spatiotemporal solar radiation forecasting: ground-based all-sky imagers, satellite-based forecasts, and NWP \cite{Kleissl2013}. Physics-based methods have been widely applied and validated for surface solar irradiance forecasting, with lead times extending up to 15 days \cite{Lorenz2012, Perez2013, Aguiar2016, Lorenz2016, Haupt2018}. These approaches typically leverage NWP models, which make use of data assimilation techniques for defining the initial conditions and integrate the Navier-Stokes equations in time to produce forecasts \cite{Zhao2016, Roberts2018}. NWP models can be tailored for specific regions or used to generate predictions at a global scale \cite{Mathiesen2011}. Drawbacks of NWP models are their relatively coarse spatial resolution and low update frequency, and their high demand for computational resources and processing time. Consequently, NWP systems like the high-resolution integrated forecast system (IFS-ENS) model developed by the European Centre for Medium-Range Weather Forecasts (ECMWF) can only generate four forecasts per day.

Geostationary satellites capture images of the Earth multiple times per hour, enabling continuous observation of any region of interest between 65$^\circ$N and 65$^\circ$S. By comparing successive satellite images, techniques such as the block matching algorithm or the Lucas-Kanade method can be adopted to estimate cloud motion vectors (CMVs) \cite{Lorenz2012, Lucas1981}. The latter describes the velocity and direction at every pixel to advect cloud cover in time. By adding perturbations to the CMVs, a forecast ensemble can be obtained that can be used to characterize forecast uncertainties \cite{Carrière2021}. Forecast models that rely on CMVs have been shown to outperform NWP models for time horizons of up to several hours \cite{Wang2019}. One major drawback of these models is their inability to capture the evolution of clouds over time \cite{Blanc2017, Urbich2018}. To address this limitation, \cite{Carpentieri2023} proposed a probabilistic CMV-based nowcasting model for the clear-sky index (CSI), i.e. the ratio of all-sky SSI to clear-sky SSI, that captures both cloud advection and cloud evolution, outperforming advection-only models by 9\% in terms of the continuous ranked probability score (CRPS). We note that sky cameras also provide cloud information at high temporal resolution, but their coverage is limited to just a few square kilometers \cite{Paletta2021, Niu2025}. As a result, they are less suitable for spatiotemporal solar forecasting across country-scale regions.

Multi-annual satellite data records have enabled the development of machine-learning (ML) driven spatiotemporal forecast models in recent years. Early ML approaches leveraged various deep learning architectures, such as long short-term memory (LSTM) networks, convolutional neural networks (CNNs), and graph neural networks (GNNs), and were trained using a combination of ground-based measurements and satellite data \cite{Lago2018, Brahma2020}. However, their dependence on ground measurements restricted their applicability to regions where such data were available. A notable advancement came with the work of \cite{Nielsen2021}, who introduced one of the first deep learning solar forecast models to rely on satellite observations. Their approach consisted of an autoencoder architecture incorporating three convolutional LSTM (ConvLSTM) layers to forecast the CSI probability at each pixel up to 4 hours ahead. Recently, \cite{Carpentieri2025} proposed a deep generative diffusion model for probabilistic spatiotemporal nowcasting of CSI. Their model combined a variational autoencoder (VAE) for dimensionality reduction with a latent diffusion model conditioned on a deterministic latent nowcaster, achieving improved forecast accuracy and a 60\% increase in extreme event prediction performance compared to the model by \cite{Nielsen2021}. 

Assessing the performance and generalizability of spatiotemporal SSI and PV power forecast models remains challenging, as it requires access to extensive networks of PV systems or ground-based SSI measurements distributed across large geographic areas. However, operational PV power datasets are scarce and, when available, usually include data from only a limited number of PV systems. Consequently, previous studies evaluating the accuracy of PV power forecast models are based on relatively small datasets, typically including at most a few dozen PV systems \cite{Aldahidi2025, Vandermeer2018, Xiang2024, Diazbello2025, Qin2022, Agoua2019, Brester2023, Perera2024, Wang2025, Ali2025, Lin2016, Markovics2022}. In terms of the number of PV systems involved, an exception is the work by \cite{Karimi2021} who trained a GNN using data from 316 PV systems, representing, to the best of our knowledge, the largest operational dataset used for this task to date.

In this work, we present the first demonstration of spatiotemporal PV forecasting, validating the reliability, sharpness, and overall forecast performance. Our comparison includes the probabilistic optical flow model SolarSTEPS \cite{Carpentieri2023}, the probabilistic deep generative diffusion model SHADECast \cite{Carpentieri2025}, the deterministic deep learning model IrradianceNet \cite{Nielsen2021}, and the ECMWF IFS-ENS model, an advanced NWP model frequently in use for SSI forecasts in practice \cite{Yang2022b, Wang2022, Sebastianelli2024}. First, we generate SSI forecasts with each model and compare them against satellite observations. After converting irradiance to PV power, we benchmark the resulting forecasts using two years of operational data from 6434 PV systems across Switzerland.

The main innovations of this study are summarized as follows: (1) we introduce the first probabilistic satellite-based spatiotemporal PV forecast approach, (2) we characterise the proposed spatiotemporal PV forecast approach on a countrywide network of more than 6400 PV systems, specifically comparing four satellite-based PV nowcasting models with the IFS-ENS model and characterising how their forecast errors depend on lead time, altitude, and cloud conditions. We also provide (3) the first visualisation of how mesoscale cloud systems impact national PV production at intra-day and intra-hour time scales.

The structure of this article is as follows. Section \ref{sec:data} describes the satellite-derived SSI dataset, the PV power dataset, and the metrics employed to evaluate forecast skill. Next, Section \ref{sec:models} describes the forecast models evaluated in this study together with the irradiance-to-power conversion model. Section \ref{sec:workflow} details the methodology adopted to perform intraday forecasts with each model. In section \ref{sec:results}, we discuss our results, which include the accuracy of the irradiance-to-power conversion models and, in particular, the SSI and PV power forecast quality of all models. We summarise our conclusions in Section \ref{sec:conclusions}.

\begin{figure}[t]
	\centering
	\includegraphics[width=1\textwidth]{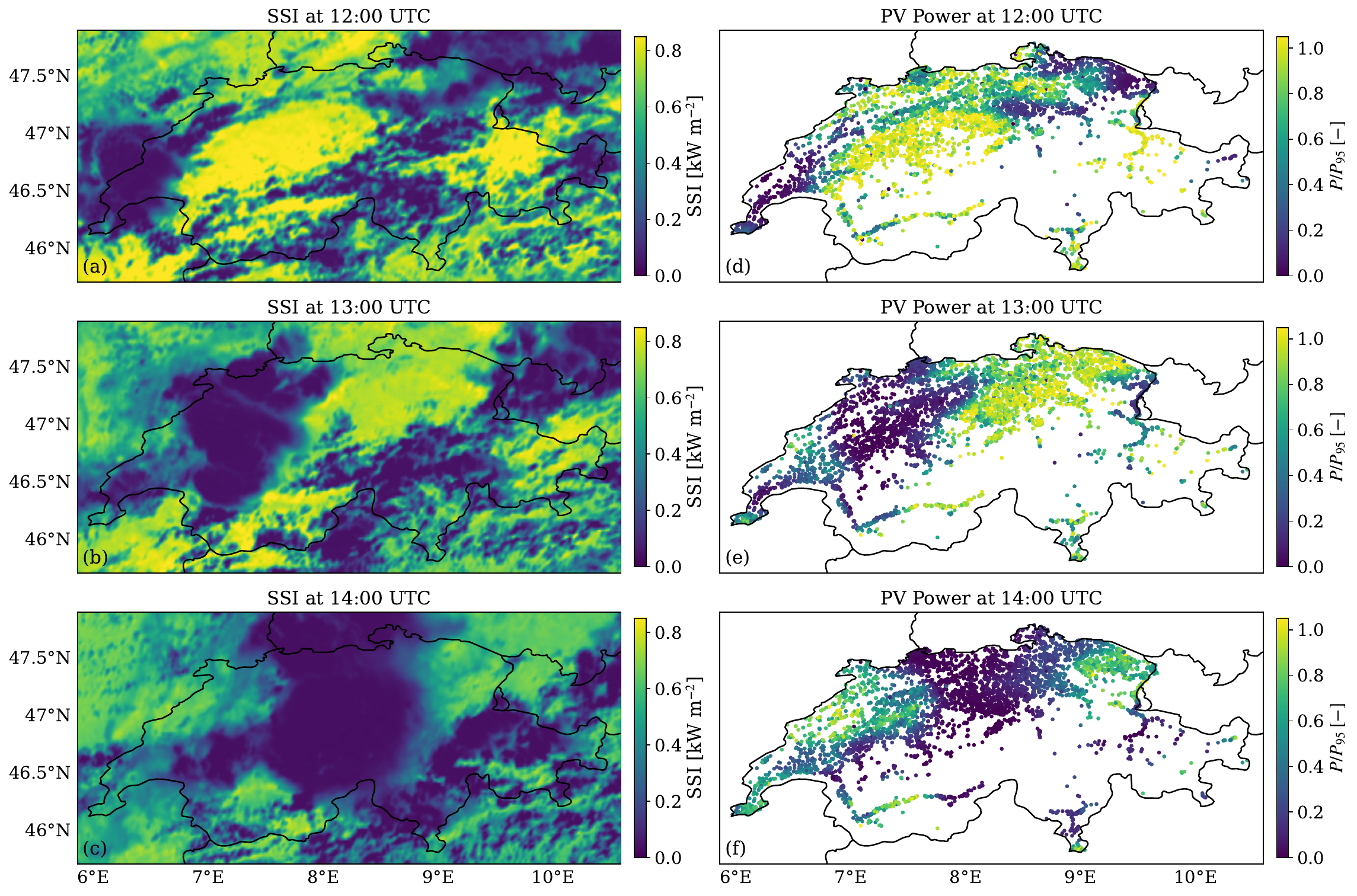}
	\caption{(a-c) Satellite-based SSI fields and (d-f) PV power output of the 6434 PV stations, observed on 6 August 2019 at three time stamps: 12:00 UTC, 13:00 UTC and 14:00 UTC. The black lines denote national borders. Note that the PV power output is normalized using the station-specific 95th percentile of the power time series.}
	\label{fig:ssr_vs_pvdata}
\end{figure}

\section{Datasets and metrics}\label{sec:data}

\subsection{Surface solar irradiance}\label{sec:data_hanna}
SSI represents the combination of direct and diffuse solar radiation incident on a horizontal plane at the Earth’s surface. It is also called global horizontal irradiance and is typically measured by ground-based pyranometers or retrieved from satellite observations. In this work, we use satellite-derived SSI fields provided by EUMETSAT Climate Monitoring Satellite Application Facility (CM SAF), notably the High-Resolution European Surface Solar Radiation Data Record (HANNA) -- see \ref{app:hanna_sarah_ssi}. The SSI values are derived from data collected by the Spinning Enhanced Visible and Infrared Imager (SEVIRI) onboard the Meteosat Second Generation (MSG) satellite positioned in geostationary orbit at 0$^\circ$ longitude \cite{Schmetz2002}, which provides Earth scans with a sampling distance at nadir of 3 km. HANNA builds on the HelioMont algorithm \citep{Stockli2013, Castelli2014} to estimate SSI from the SEVIRI channel intensities. The resulting satellite data are projected onto a grid with a spatial resolution of 0.01$^\circ$ for both longitude and latitude, covering a domain of [-17$^\circ$W, 35$^\circ$E] $\times$ [30$^\circ$N, 70$^\circ$N]. The MSG SSI fields are available for the years 2019–2020 at a temporal resolution of 15 minutes \cite{Hanna2025}.

The HANNA SSI has undergone initial validation using ground-based measurements from several networks, including the Baseline Surface Radiation Network (BSRN), the Global Energy Balance Archive (GEBA), Deutscher Wetterdienst (DWD), and SwissMetNet. In total, 406 stations were used for this validation. Results indicate that, for stations located in the 0–500 meter elevation band, the station- and monthly-averaged mean absolute difference measures 5.3 W m$^{-2}$ while the same error measures 6.9 W m$^{-2}$ and 10.9 W m$^{-2}$ for stations lying in the elevation range 500-1500 and above 1500 m, respectively \cite{Hanna2025}. It is worth noting that 5 out of the 731 days in the dataset were excluded from our analysis due to the presence of missing values for some of the time steps over our region of interest. This issue is also acknowledged in \cite{Hanna2025}.

Figure \ref{fig:ssr_vs_pvdata}(a–c) displays HANNA SSI fields obtained over the study area on the 6th of August 2019 at three time stamps: 12:00, 13:00, and 14:00 UTC. These panels illustrate how the dataset enables a sharp representation of cloud structures (panels (a–c)), identifiable by their radiative effect on the surface as areas with lower SSI values. During this period, we also observe convective cloud growth over central Switzerland, along with a gradual decrease in average SSI consistent with the progressing diurnal cycle. In \ref{app:ssi_power_forecast}, the full two-hour period at 15-minute temporal resolution is shown. 

\begin{figure}[t]
	\centering
	\includegraphics[width=1\textwidth]{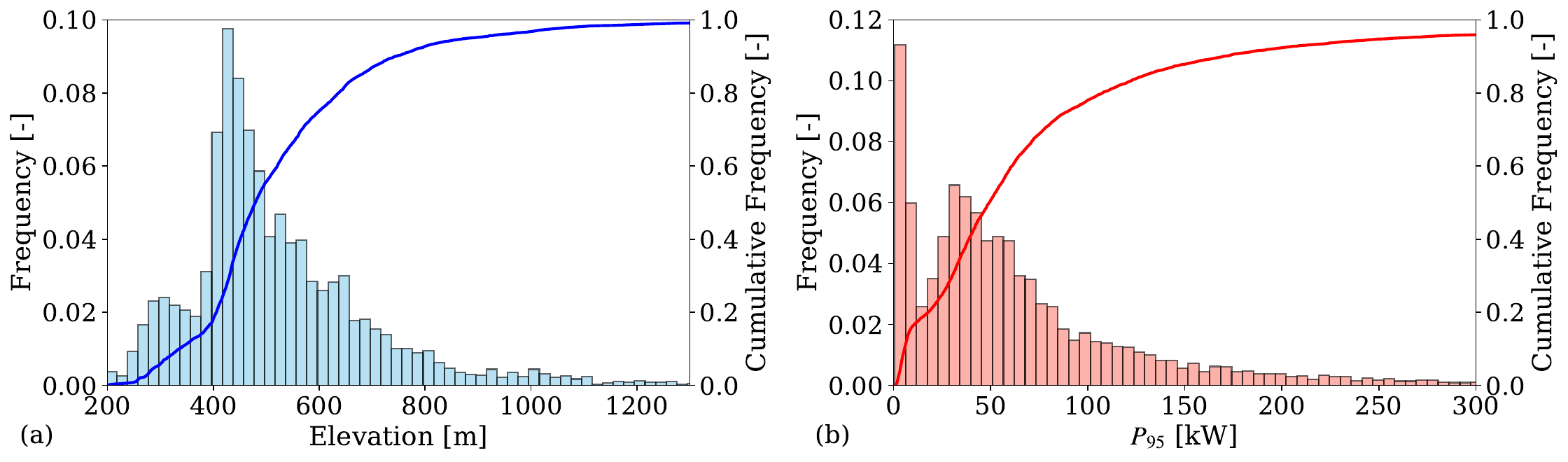}
	\caption{Empirical probability density function and cumulative density function (CDF) of the PV stations (a) elevation and (b) $P_{95}$ value.}
	\label{fig:pvdata_distribution}
\end{figure}

\subsection{PV power production}\label{sec:pvdata}
Our study focuses on the years 2019–2020, which coincide with the HANNA SSI record. This resulted in 7633 usable PV stations from across Switzerland with 15-minute resolution measurements, whose proprietary production data were obtained for this study under a data usage agreement. Given the large multi-annual record of thousands of PV stations, we developed and applied automated cleaning procedures to identify and filter anomalous data to ensure data quality. Specifically, we split the PV production time series of each PV station into two segments, one for 2019 and one for 2020, and computed statistical metrics such as mean, standard deviation, and skewness for each year. Given the seasonal stability in the statistical properties of solar generation, we expected these distributions to be broadly similar. We therefore applied a 10\% tolerance threshold. If any of the computed metrics differed by more than this threshold between the two years, the PV station was flagged as an outlier and excluded from the dataset. This filtering strategy reduced the number of stations considered in this study from 7633 to 6434.  Assuming an average of 12 hours of daylight per day over the year, our study relies on roughly 225 million PV production data points that can be used to train or validate forecast models. In the remainder of the article, the PV power generated from each station is normalized with $P_{95}$, i.e. the station-specific 95th percentile of the power time series. This normalization provides a robust basis for comparing stations with different capacities, mitigates the impact of rare extreme values while still preserving the upper range of realistic operating conditions.

Figure \ref{fig:ssr_vs_pvdata}(d-f) shows the location and power output of the 6434 PV stations considered in this study, measured on the 6th of August 2019 at three time stamps: 12:00, 13:00, and 14:00 UTC. The cloud system over Switzerland, visible in the SSI fields shown in Figure \ref{fig:ssr_vs_pvdata}(a-c), is clearly reflected in the PV power output. Areas of low PV power generation align remarkably well with regions of low SSI, demonstrating a strong correlation between these variables. We note that the full two-hour period at 15-minute temporal resolution is shown in \ref{app:ssi_power_forecast}.  Figure \ref{fig:ssr_vs_pvdata} highlights the critical importance of spatiotemporal information over purely time-series-based methods unaware of the surroundings SSI field for forecasting solar energy. To our knowledge, it is the first visualisation of how mesoscale cloud systems affect the production of a large country-wide fleet of PV systems. It illustrates how, by relying on geostationary satellites, we can infer the footprint of cloud cover on the country-wide PV production with high accuracy and low latency. 

Figure \ref{fig:pvdata_distribution}(a) shows the distribution of the site elevation of the 6434 PV systems analyzed in this study. Approximately 56\% of the stations are situated between 0–500 m above sea level (a.s.l.), a range where the HANNA dataset exhibits the highest accuracy. Fewer than 1\% of the stations are located above 1300 m a.s.l., with the highest reaching an elevation of approximately 2450 m a.s.l. Figure \ref{fig:pvdata_distribution}(b) illustrates the distribution of the 95th percentile of the power time series of each station and its corresponding cumulative density function (CDF). About 16\% of the stations have a $P_{95}$ value below 10 kW, typically corresponding to residential installations. The majority of the stations (about 76\%) fall within the 10–200 kW range while only 4\% have a $P_{95}$ value exceeding 300 kW.

\begin{figure}[t]
	\centering
	\includegraphics[width=1\textwidth]{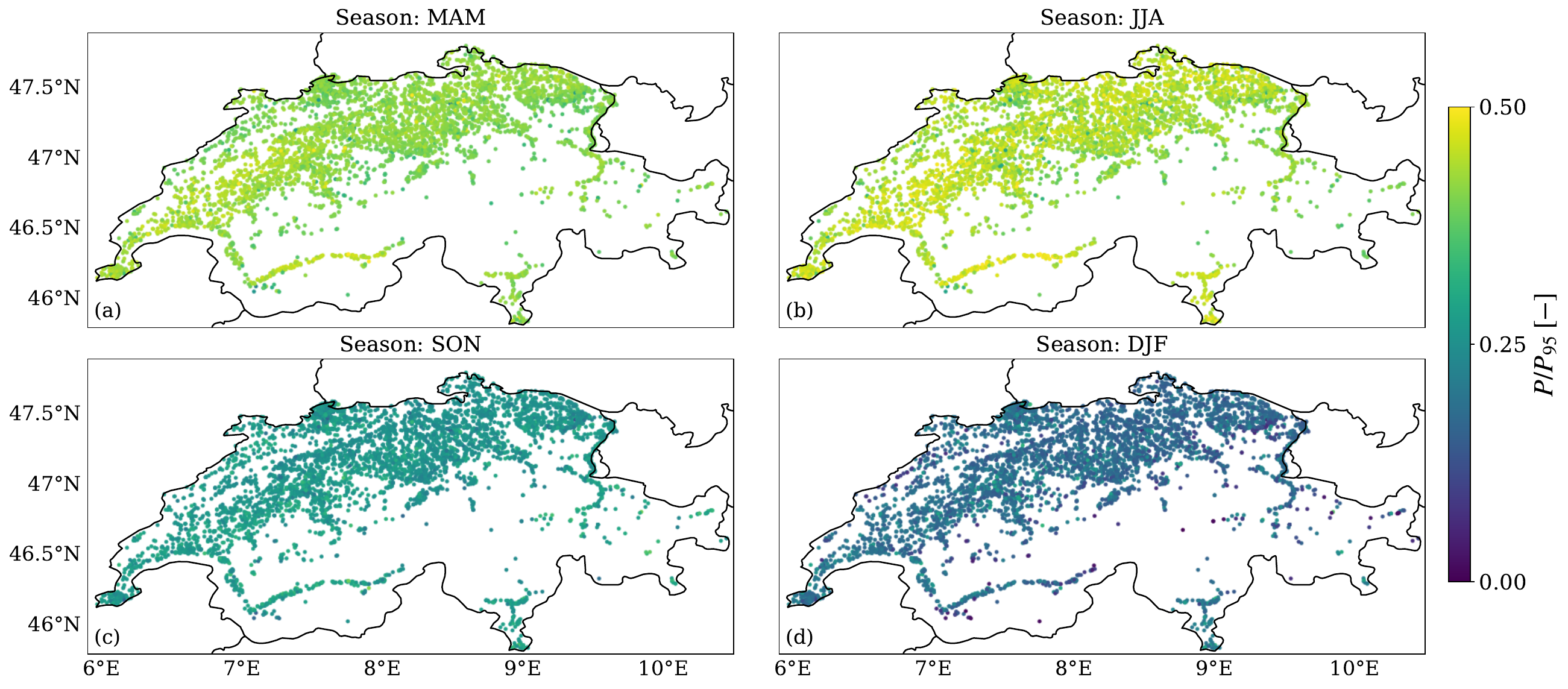}
	\caption{Two-year average of PV power output normalized with the station-specific $P_{95}$ value during (a) MAM, (b) JJA, (c) SON, and (d) DJF. The averaging is performed over timestamps between local sunrise and sunset times.}
	\label{fig:pvdata_mean_power}
\end{figure}

Finally, Figure \ref{fig:pvdata_mean_power} shows the seasonally averaged normalized power output calculated during daytime hours (from sunrise to sunset) over the 2019–2020 period. While the intraseasonal variability is relatively low, interseasonal differences are significant as expected. For instance, the station-averaged $P/P_{95}$ ratio reaches 0.43 in summer (JJA), but drops to 0.18 in winter (DJF).

\subsection{Metrics}\label{sec:data_metric}
To evaluate the performance of the forecast models, we employ different metrics for deterministic and probabilistic predictions. For deterministic forecasts, accuracy is assessed using the mean absolute error (MAE), root mean square error (RMSE), and mean bias error (MBE). The MAE provides an intuitive measure of the average magnitude of the forecast errors, while the RMSE assigns greater weight to large deviations and is therefore more sensitive to extreme errors. The MBE, on the other hand, captures the average signed error, thereby indicating whether a model tends to systematically overestimate or underestimate the observations. 

For probabilistic forecasts, deterministic scores are computed using the ensemble mean, while additional metrics evaluate both the reliability and sharpness of the predictive distributions. The prediction interval coverage probability (PICP) measures the proportion of observations that fall within the forecast intervals and is thus an indicator of calibration. Complementary to this, the mean prediction interval width (MPIW) and the prediction interval normalized average width (PINAW) quantify the average width of the prediction intervals and reflect the forecast sharpness, with narrower intervals being preferable provided that coverage remains adequate. The continuous ranked probability score (CRPS) offers a more comprehensive evaluation by integrating information about both reliability and sharpness into a single metric. For deterministic models, the CRPS reduces to the MAE \cite{Hersbach2000}. Finally, rank histograms provide a graphical diagnostic of ensemble forecasts, revealing whether the ensemble spread appropriately represents the observations and highlighting potential issues such as bias, underdispersion, or overdispersion. We note that, when the metric acronym is preceded by the letter 'n', this indicates that the metric has been normalized.

Taken together, these metrics allow for an in-depth comparison of deterministic and probabilistic models, capturing not only point prediction accuracy but also the quality of uncertainty quantification. For more details on the definition and computation of these metrics as well as the normalization factors, we refer the reader to \ref{app:metrics}.

\section{Models}\label{sec:models}
To enable a comprehensive comparison between satellite-based spatiotemporal nowcasting models and a physics-based NWP model, we consider seven distinct approaches, covering both deterministic and probabilistic methods. Section \ref{sec:models_evaluated} provides a brief overview of each model and a summary of its underlying assumptions and design choices. Finally, Section \ref{sec:ssitopower} describes the methodology adopted to convert SSI into PV power.

\subsection{SSI forecast models}\label{sec:models_evaluated}

\subsubsection{SolarSTEPS}\label{sec:solarsteps}
SolarSTEPS is a probabilistic optical-flow-based model developed for forecasting satellite-derived CSI fields \cite{Carpentieri2023}. First, the model computes CMVs from a sequence of CSI fields using the Lucas-Kanade optical-flow algorithm \cite{Lucas1981}. Subsequently, a Fast Fourier Transform (FFT) is applied to decompose each input CSI field into distinct spatial scales. Each spatial scale is then forecast using a separate linear autoregressive (AR) model. These AR models are designed to simulate the temporal evolution of cloud patterns in a Lagrangian frame and include spatially correlated noise to generate an ensemble of forecasts. Each scale is modelled with its own set of AR coefficients. The forecast components from all spatial scales are then summed and advected using the initially estimated CMVs to produce the final forecast.

SolarSTEPS stands out for its ability to capture both cloud advection, using an optical-flow algorithm, and cloud evolution, through scale decomposition combined with AR models. Additionally, it supports the generation of ensemble forecasts, allowing for uncertainty quantification. However, as linear AR models assume data stationarity, the model is unable to predict distribution shift, which occurs in weather conditions in which cloudiness is growing or decreasing over the course of the forecast. Additionally, we also evaluate a simplified version of SolarSTEPS, referred to as SolarSTEPS-pa. This pure advection (PA) variant does not model the temporal variability of the CSI. Instead, it perturbs the CMVs derived from the input image sequence and uses them to advect the CSI. As a result, this version captures only the effect of cloud advection, omitting both the scale decomposition and AR modelling components.
The model was originally calibrated on the HelioMont dataset \cite{Castelli2014}, and the AR models coefficients were estimated using the Yule-Walker equations \cite{Brockwell1991}. Moreover, the ensemble forecast consists of 10 members. In this work, we adopt the same parameter settings as reported in \cite{Carpentieri2023} to which we refer for more details.

\subsubsection{IrradianceNet}\label{sec:irradiancenet}
IrradianceNet is a deterministic deep learning model introduced by \cite{Nielsen2021}, designed to forecast satellite-derived CSI fields using only satellite-based CSI inputs. The model employs a spatiotemporal autoencoder architecture composed of three ConvLSTM layers in both the encoder and decoder. The encoder processes sequences of CSI fields to capture spatiotemporal dependencies. The final hidden state of the encoder serves as a compressed representation of the input dynamics, which is then passed to the decoder. The decoder performs upsampling to generate future CSI fields, effectively achieving spatiotemporal forecasting.

IrradianceNet was one of the first deep learning approaches for solar irradiance forecasting and has shown significant agreement with ground-based pyranometer observations \cite{Nielsen2021}. In addition to forecast blurring, a key limitation of the model is its deterministic nature, which prevents ensemble forecasting and therefore does not allow for uncertainty quantification. However, we include it in our study since it provides a reliable deterministic baseline.

The model was originally trained on the SARAH-2.1 dataset provided by EUMETSAT \cite{Pfeifroth2019}. However, \cite{Carpentieri2023} later retrained it using HelioMont data \cite{Castelli2014}, extending the forecast horizon from 2 to 8 time steps via an autoregressive strategy. In this work, we adopt the same model architecture and pre-trained weights as presented in \cite{Carpentieri2023}.

\subsubsection{SHADECast}\label{sec:shadecast}
SHADECast is the first deep generative diffusion model developed for intraday solar energy forecasting. The model takes as input a sequence of CSI fields and combines a variational autoencoder (VAE), a latent deterministic nowcaster, and a latent diffusion model to generate probabilistic SSI forecasts. The VAE, built with a symmetric architecture of two downsampling 3D residual blocks, compresses the input CSI sequence into a latent representation. This latent representation is then passed to a deterministic nowcaster operating in latent space, which comprises four Adaptive Fourier Neural Operator (AFNO) blocks, followed by a temporal transformer and an additional four AFNO blocks. To generate the probabilistic forecast, a latent diffusion model is employed, which maps Gaussian noise to future CSI representations. The diffusion process is guided by a latent denoiser with a symmetric U-Net architecture, conditioned on the output of the nowcaster. Finally, the VAE decoder transforms the forecast latent representations back into the physical space, generating the predicted CSI fields.

As a state-of-the-art model for solar energy nowcasting, SHADECast is included in our study. The model was trained on satellite-derived CSI fields from the HelioMont dataset and generates ensemble forecasts with 10 members. In this work, we adopt the same architecture and pre-trained weights as described in \cite{Carpentieri2025}, and refer to that source for detailed information on the model architecture and training procedure.

\subsubsection{IFS-ENS}\label{sec:hres}
We incorporate forecasts from the IFS-ENS model developed by ECMWF \cite{Roberts2018} to also include a physics-based probabilistic forecast model. The model provides SSI forecasts four times a day at 00:00, 06:00, 12:00, and 18:00 UTC, with outputs available at hourly intervals. To maintain consistency with the probabilistic satellite-based models, we randomly select 10 ensemble members from the original set of 50. The IFS-ENS forecasts offer a horizontal resolution of approximately 9 km, corresponding to a spatial resolution of about 0.08$^\circ$ over the area of interest.

To investigate whether bias correction could improve performance, we applied an ML–based correction to the IFS-ENS SSI forecasts. The correction network is a symmetric U-Net–based convolutional neural network \cite{Ronneberger2015}. Further details on the training procedure and network design are provided in \ref{app:bias_correction}. Throughout this work, we refer to the bias-corrected forecasts as IFS-ENS-corrected.

\subsubsection{Persistence}\label{sec:persitence}
The persistence model is a widely used benchmark for intra-hour and intra-day SSI forecasting. It assumes that future irradiance values will remain equal to the most recent observation \cite{Dutta2017}. Persistence could perform well for very short lead times due to the smooth temporal evolution of cloud fields, but its skill deteriorates rapidly for longer horizons or under rapidly changing cloud conditions. In this work, we include the persistence model to establish a clear and interpretable benchmark.

\subsection{Irradiance-to-power conversion}\label{sec:ssitopower}
To enable validation on the PV power production dataset presented in Section \ref{sec:pvdata}, the satellite-based and NWP models SSI forecasts have to be converted to corresponding PV power output. The relationship between SSI and PV power is often referred to as the solar power curve \cite{Yang2022, Mayer2022}. Several physical and semi-empirical models have been developed for this purpose. They typically comprise a chain of sub-models, including separation \cite{Yang2021}, transposition \cite{Reindl1990}, temperature \cite{Jimenez2016}, inverter \cite{Chu2021}, shading \cite{Mayer2020}, and loss models \cite{Nguyen2016}, and require detailed technical specifications of each PV installation, such as panel orientation, tilt angle and module electrical characteristics \cite{Li2025}. However, obtaining such detailed information is highly challenging when working with thousands of PV installations. Furthermore, since only SSI observations are available, estimating the direct and diffuse components would require additional modelling steps, thereby introducing further uncertainties and potential biases into the model chain. For these reasons, we adopt a data-driven regression approach that maps the SSI forecasted values to PV power production in a site-specific manner. The main limitation of a data-driven approach lies in its dependence on historical time-series data, making it unsuitable for PV systems without any prior production records. However, given the extensive availability of operational PV power data in our dataset, an ML-based conversion approach is well supported. 

\begin{figure}[t]
	\centering
	\includegraphics[width=1\textwidth]{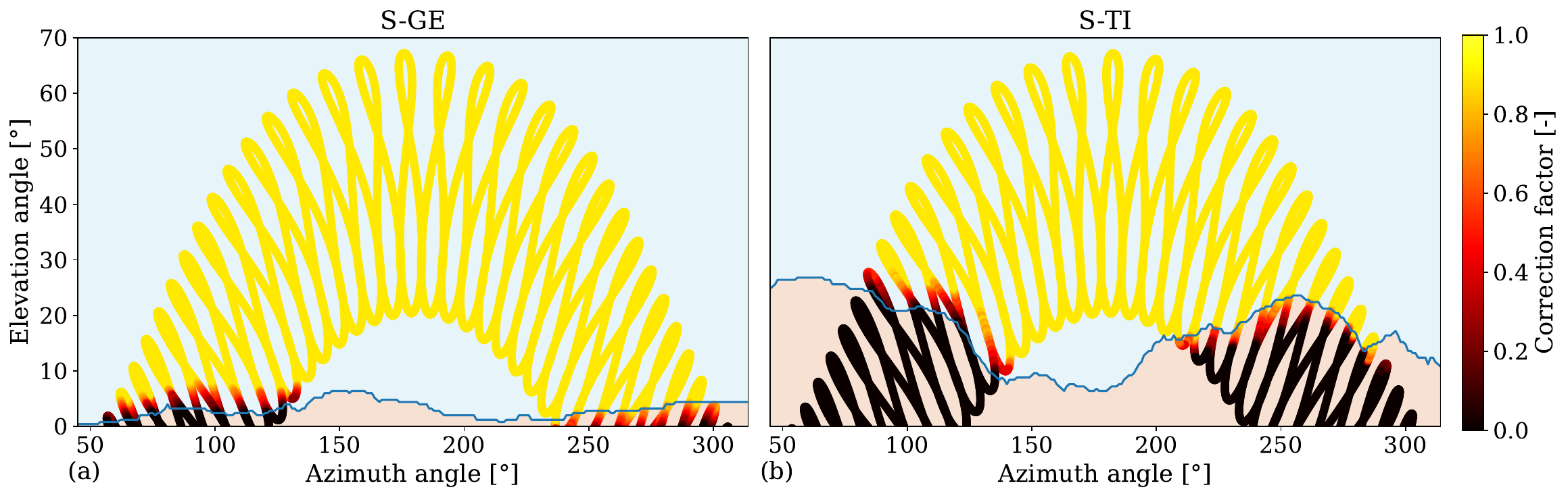}
	\caption{Solar zenith and elevation angles computed over a full year at 30-minute resolution for two PV systems located in the vicinity of (a) Geneva and (b) Biasca. The correction factor represents the fraction of time since the previous timestamp during which the sun remains above the local horizon, ranging from 0 (entirely below the horizon) to 1 (entirely above the horizon). This figure was generated using the HORAYZON library \cite{Steger2022}.}
	\label{fig:horayzon}
\end{figure}

The power output of a PV system is influenced by both meteorological and geographical factors. The most significant meteorological predictor is the SSI, showing a strong correlation with PV power output. For instance, \cite{Ahmed2020} reported an R-squared coefficient of 0.98. Other factors, such as air temperature, cloud type, dew point, and relative humidity, show progressively lower correlations, with wind speed being the least relevant \cite{Ahmed2020}. In our framework, we selected SSI as the sole meteorological predictor. Additionally, we incorporated the solar zenith angle (SZA) and the solar azimuth angle (AZI) to account for geographical effects, such as shading by mountains. SZA and AZI were computed using HORAYZON, a ray-tracing algorithm for computing the local horizon and sky view factor \cite{Steger2022}. This is particularly important in regions where topographical features can cause substantial variation in SSI. Figure \ref{fig:horayzon}(a,b) illustrates the importance of accounting for local SZA and AZI angles by showing the sun position over an entire year at 30-minute intervals for two PV installations in the vicinity of Geneva and Biasca, two locations in Switzerland, which we will denote with S-GE and S-TI, respectively. Notably, the S-TI site receives considerably less direct sunlight during the winter months. Without accounting for the local SZA and horizon, this reduced irradiance could mistakenly be attributed to cloud cover. 

To capture the periodicity of time, we also included both the day of the year (DoY) and the hour of the day (HoD) as predictors. Instead of using their raw values, we applied sine and cosine transformations to preserve their cyclical nature, therefore introducing four additional predictors. In summary, each PV system is characterized by seven predictors, i.e. SSI, SZA, AZI and the four time-based features derived from the DoY and HoD, and one target variable, i.e. its measured PV power output.

We perform the irradiance-to-power conversion with the extreme gradient boosting (XGBoost) algorithm \cite{Chen2016} due to its good performance and compatibility with GPU acceleration. We note that alternative ML models, such as decision trees \cite{Quinlan1986}, support vector machines \cite{Smola2004}, light gradient boosting \cite{Ke2017}, histogram-based gradient boosting \cite{Scikitlearn2011}, and long short-term memory networks \cite{Hochreiter1997}, were also explored, but they did not yield performance improvements over XGBoost for this task. To account for differences in nominal capacity and elevation among the PV systems considered in this study, we trained a separate XGBoost model for each station. Each model uses the HANNA SSI fields interpolated to the station location together with the corresponding operational PV power measurements, using the predictors and target variable described above. The two-year dataset was partitioned into non-overlapping blocks of 12 days. Within each block, ten days were assigned to the training set, while the remaining two days were used for validation and testing. Only daylight periods, i.e. between sunrise and sunset, were included. All input and target variables were scaled to the [0,1] range to improve numerical stability and facilitate model convergence. The hyperparameters of each model, such as the learning rate, maximum tree depth, number of estimators, and the L1 and L2 regularization terms, were tuned independently for each station using Optuna \cite{Optuna2019}. In total, we trained 6434 station-specific irradiance-to-power conversion models. Their performance on the test set is presented in Section~\ref{sec:result_ssi_to_power}.

\section{Spatiotemporal PV forecast workflow}\label{sec:workflow}
The satellite-based nowcasting models, i.e. IrradianceNet, SolarSTEPS, and SHADECast, were originally trained and calibrated on seven years of satellite-derived CSI fields from the HelioMont dataset at a spatial resolution of 0.02$^\circ$. For our evaluation, we use two years of HANNA data, which provides SSI at a resolution of 0.01$^\circ$. To maintain consistency with the training setup, the HANNA fields are downsampled to 0.02$^\circ$ using a 2$\times$2 pixel averaging kernel. SSI fields are then converted to CSI, a dimensionless ratio of observed SSI to clear-sky SSI (SSI$_\mathrm{cs}$) that typically ranges from 0 to approximately 1.2. SSI$_\mathrm{cs}$ is computed using the Ineichen model \cite{Ineichen2008} based on precomputed look-up tables from the SOLIS model \cite{Mueller2004}. This method was chosen for its independence from external data and is implemented in the pvlib Python package \cite{pvlib2018}. Each nowcast takes four consecutive CSI fields (covering one hour) as input and predicts the next eight fields, corresponding to a two-hour forecast horizon, which are subsequently converted back to SSI. Given HANNA's 15-minute temporal resolution, this input-output sequence spans 3 hours in total.

SHADECast and IrradianceNet operate on upscaled CSI fields from HANNA over the region [4.41$^\circ$E, 12.09$^\circ$E] $\times$ [42.99$^\circ$N, 50.67$^\circ$N], corresponding to images of size 384 $\times$ 384 pixels. This domain covers a region centred around Switzerland, which consists of our area of interest. Given this domain size, SHADECast requires 6.5 seconds to generate an ensemble forecast with 10 members on a single NVIDIA GH200 GPU. Due to architectural constraints, IrradianceNet cannot process arbitrarily large inputs. Therefore, each CSI field is divided into nine patches of size 128 x 128 pixels, with linear interpolation applied along the borders of each patch to reconstruct the full field. This patching strategy, previously adopted by \cite{Nielsen2021, Carpentieri2025}, allows IrradianceNet to be applied to arbitrarily large spatial domains. Consequently, a single SHADECast inference corresponds to nine separate inferences with IrradianceNet. The latter requires approximately 1.9 seconds to produce a deterministic forecast over a 128 × 128 pixel area on the same GPU model, resulting in a total inference time of about 17 seconds for the full domain.

SolarSTEPS employs optical-flow techniques to generate CSI forecasts, which can result in regions with missing values. This is an inherent limitation of optical-flow models, as they advect CSI values using estimated CMVs but do not extrapolate new information beyond the spatial coverage of the input data. As a result, areas advected from regions lacking upstream data remain undefined in the forecast outputs \cite{Carpentieri2023}. To mitigate this issue, SolarSTEPS and its simplified variant SolarSTEPS-pa use slightly larger input domains, defined over the region [3.13$^\circ$E, 13.37$^\circ$E] $\times$ [42.99$^\circ$N, 50.67$^\circ$N], corresponding to 512 $\times$ 384 pixels. On this input size, generating a forecast ensemble with 10 members with SolarSTEPS and SolarSTEPS-pa requires 18.0 and 6.5 seconds, respectively, on a single AMD EPYC 7742 CPU. The longer computational time attained by SolarSTEPS results from the additional calculations required to model cloud evolution.

The comparison between the satellite-based models and the NWP model IFS-ENS is performed by adopting a user-centric evaluation approach that reflects real-world forecasting needs. For example, suppose a user requests a forecast at noon and the current time is 11:55. In that case, the satellite-based models generate predictions for 12:00 to 13:45 using SSI fields observed at 11:00, 11:15, 11:30, and 11:45. For IFS-ENS, the most recent forecast covering the desired prediction window is used. Consequently, although we will refer to the forecast at 12:00 and 13:00 as having a 15- and 75-minute lead time, this will actually correspond to a different lead time in the case of IFS-ENS.

For each day, forecasts are generated for full-hour timestamps between 1 h after sunrise and 3 h before sunset. This process is repeated for the full two-year period, giving a total of 6158 inferences per model. Although satellite-driven models can generate new forecasts every 15 minutes, we restrict the evaluation to one forecast per hour to facilitate the comparison with IFS-ENS. As each forecast is independent, we leverage high-throughput computing to run them in parallel, assigning one job per processor on a compute node. The computations were performed on the Swiss high-performance computing system Alps at the Swiss National Supercomputing Centre.

\section{Results}\label{sec:results}
We begin by evaluating the accuracy of the irradiance-to-power conversion models in Section \ref{sec:result_ssi_to_power}. Next, Section \ref{sec:result_ssi} compares the SSI forecasts of all models using the HANNA SSI fields as reference. Finally, Section \ref{sec:result_power} presents the intraday PV power forecast evaluation based on operational data from the 6434 PV systems.

\begin{figure}[t]
	\centering
	\includegraphics[width=1\textwidth]{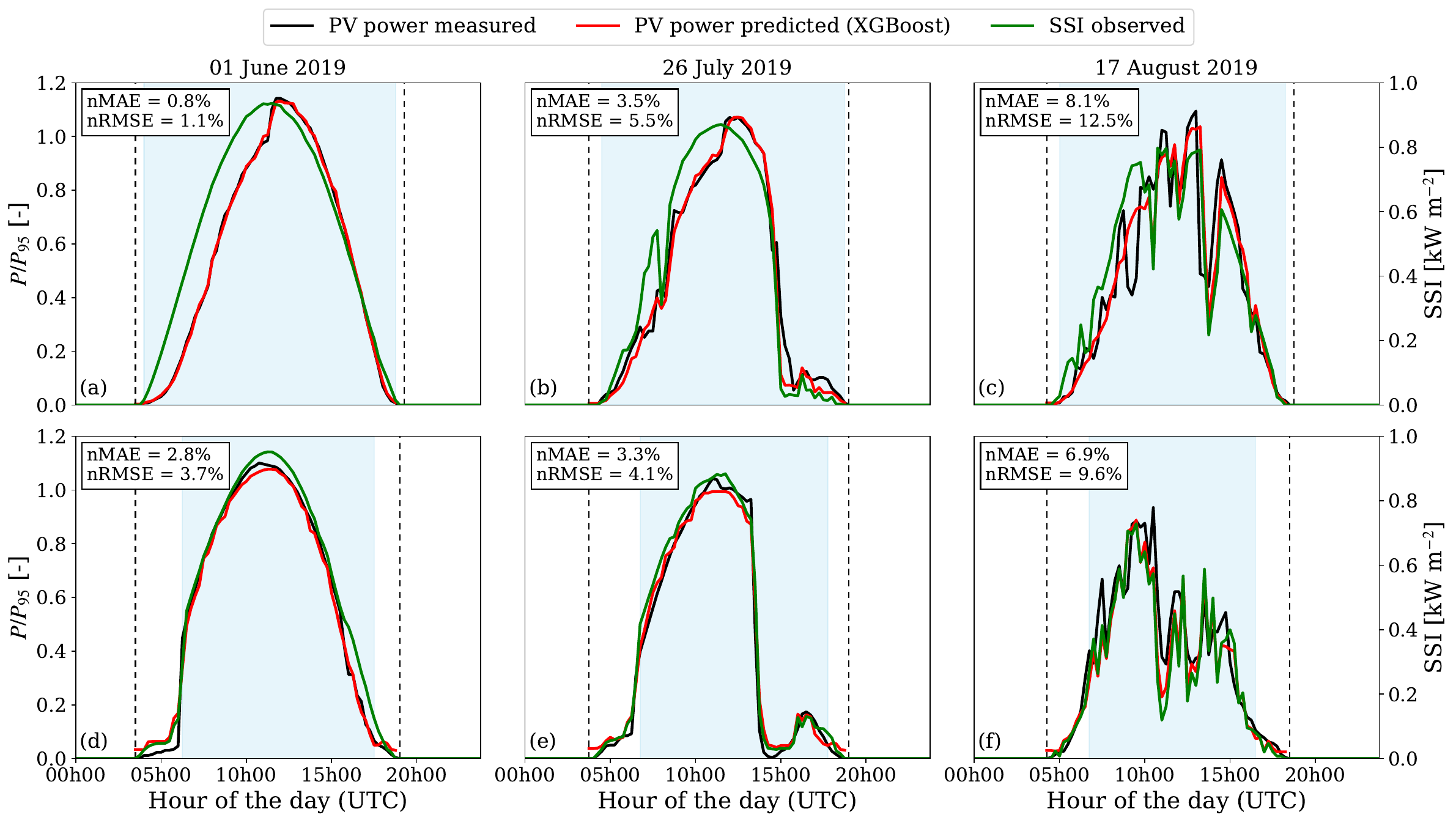}
	\caption{PV power measurements, XGBoost predictions of PV production, and SSI observations at 15-minute resolution for stations S-GE (a–c) and S-TI (d–f) over three days that belong to the test set. SSI observations from the HANNA dataset are interpolated to the corresponding station locations. The PV power values are normalized with the station-specific $P_{95}$ value. The light-blue shaded area highlights the period when the sun is above the horizon, determined from local SZA using HORAYZON. The vertical black dashed lines indicate local sunrise and sunset times. The nMAE and nRMSE values are computed for each individual day and normalized using the station-specific $P_{95}$ value.}
	\label{fig:ssi_to_power_station}
\end{figure}

\subsection{Irradiance-to-power conversion}\label{sec:result_ssi_to_power}
An initial qualitative assessment of model performance is provided by comparing measured and predicted PV power output, together with the corresponding observed SSI, for stations S-GE and S-TI over three randomly selected test-set days. Figure \ref{fig:ssi_to_power_station}(a,d) shows results for a clear-sky day, characterized by a smooth, parabolic SSI profile. The S-GE station, situated in a relatively flat region (see Figure \ref{fig:horayzon}(a)), exhibits no major impacts from the local horizon in its PV profile. In contrast, S-TI is located in a mountainous area where power output begins to ramp up only once the sun is visible above the local horizon. In both cases, the irradiance-to-power conversion model predicts the produced PV power remarkably well. Figure \ref{fig:ssi_to_power_station}(b,e) displays a day with moderate variability, where SSI ramps cause sudden drops in the power output of the two stations. The associated drops in PV power output are captured by the XGBoost model, although prediction errors are overall higher than in the clear-sky scenario. Finally, Figure \ref{fig:ssi_to_power_station}(c,f) shows a day with highly variable conditions characterized by intermittent cloud cover. Although the model captures the overall temporal evolution, the daily nMAE and nRMSE are considerably higher than in the low- and moderate variability conditions. In all cases, we find a strong positive correlation between SSI and PV power output. This highlights the importance of high-quality SSI observations and satellite retrievals for accurate irradiance-to-power conversions \cite{Schuurman2024}. We remark that the conversion is applied only between sunrise and sunset, as PV power output is zero outside this period.

\begin{figure}[t]
	\centering
	\includegraphics[width=1\textwidth]{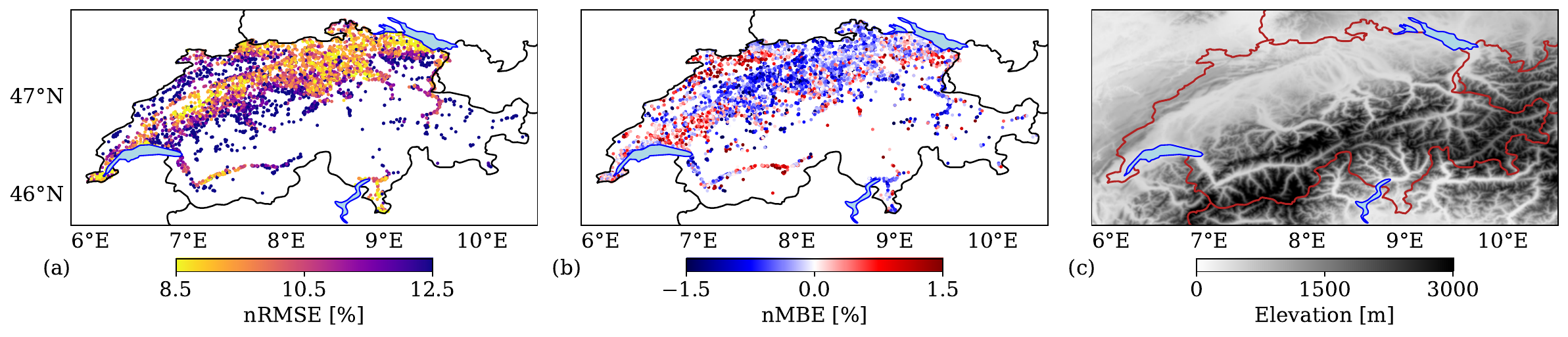}
	\caption{(a) nRMSE and (b) nMBE of the predicted PV power, averaged over the test set. Both quantities are normalized with the station-specific $P_{95}$ value. (c) Elevation map showing values in meters above sea level.}
	\label{fig:ssi_to_power_maps}
\end{figure}

\begin{figure}[t]
	\centering
	\includegraphics[width=1\textwidth]{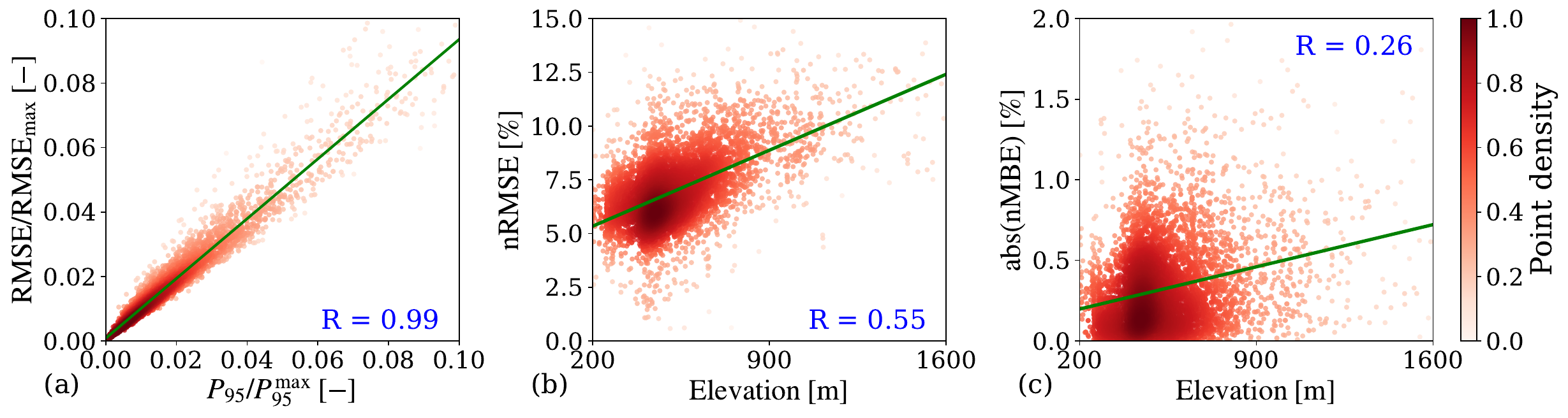}
	\caption{Scatter plot of (a) RMSE as a function of the station-specific $P_{95}$ value, with both variables normalized by their respective maximum values, and (b, c) nRMSE and nMBE as a function of elevation. The nRMSE and nMBE are averaged over the test set. The color indicates local data density estimated using a Gaussian kernel (nonlinear color normalization with $\gamma=0.3$ is applied). The green dashed line indicates a linear fit, and the Pearson correlation coefficient R is reported in blue.} 
	\label{fig:ssi_to_power_scatter}
\end{figure}

We trained a separate XGBoost model for each PV station. Figure \ref{fig:ssi_to_power_maps}(a,b) shows the nRMSE and nMBE of the predicted PV power, averaged over the test set (i.e. 61 days) and normalized by the station-specific $P_{95}$ value. Overall, nRMSE values range from 8–10\% in lowland regions, increasing to 11–13\% in mountainous areas, i.e., the Jura in the Northwest and the Alps in central and Southern Switzerland. Approximately 3.6\% of stations exhibit nRMSE values above 15\%. These stations are located at an average elevation of 873 m, compared to 514 m for the remaining 96.4\% of stations. We suspect that one of the reasons for the reduced performance of the irradiance-to-power conversion at higher elevations is the lower accuracy of HANNA SSI estimates in mountainous terrain \cite{Hanna2025}. This highlights how a dense PV monitoring network can also serve as an effective means of evaluating SSI retrieval methods. Moreover, high-elevation PV systems typically experience more variable atmospheric conditions and frequent snow cover, both of which can introduce additional discrepancies between measured and predicted power output. Figure \ref{fig:ssi_to_power_maps}(b) illustrates the nMBE. Most stations (65.7\%) exhibit a negative nMBE, suggesting that the XGBoost models tend to underestimate PV production. This bias may be related to the use of RMSE-based loss minimisation, which promotes regression toward the conditional mean and can therefore lead to a systematic underestimation of extreme PV production values. In absolute terms, however, only 10.7\% of stations show an nMBE larger than 1\%. 

\begin{figure}[t]
	\centering
	\includegraphics[width=0.65\textwidth]{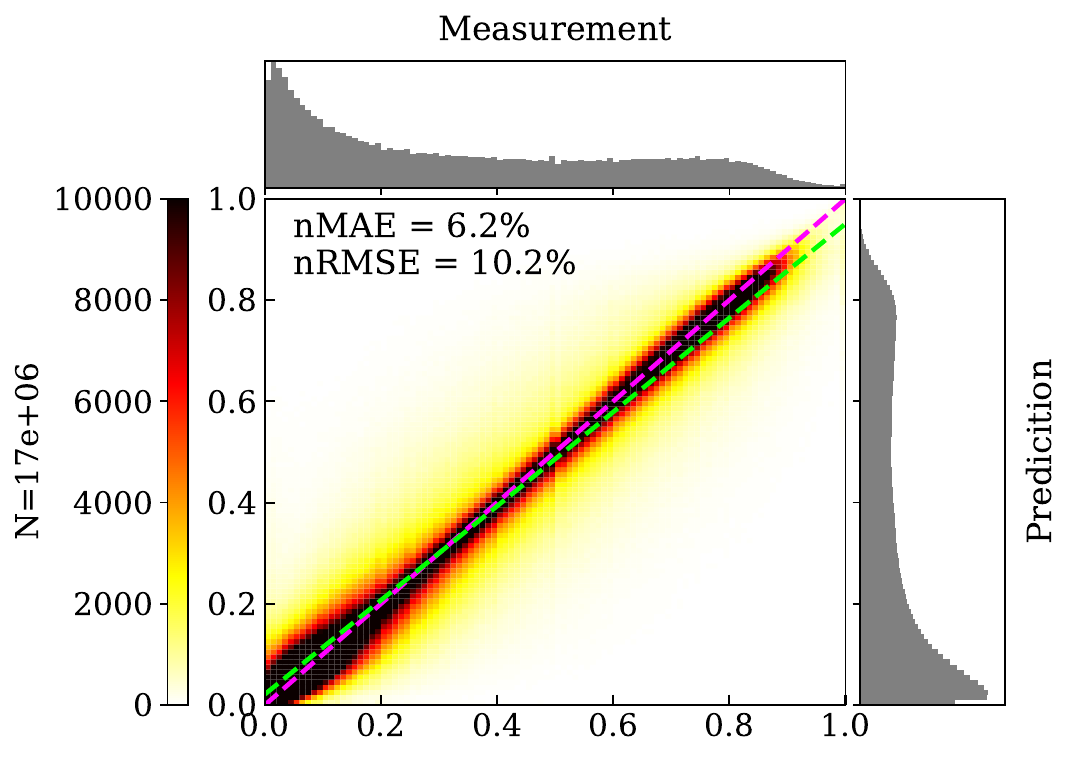}
	\caption{Scatter plot of normalized PV power measurements (x-axis) versus XGBoost model predictions (y-axis) over the test set. For each station, the values are scaled to the [0, 1] range. The pink dashed line represents the 1:1 reference, while the light green dashed line shows the linear regression fit. The nMAE and nRMSE are computed as averages across the full test set and all stations, and are normalized with the station-specific $P_{95}$ value. The marginal histograms above and at the side of the scatter plot display the distributions of measurements and predictions, respectively, using 100 bins. The plot includes approximately 17 million PV power measurements.}
	\label{fig:ssi_to_power_residual}
\end{figure}

Further analysis in Figure \ref{fig:ssi_to_power_scatter} shows how nRMSE and nMBE relate to both $P_{95}$ and elevation. As expected, the RMSE is strongly correlated with the $P_{95}$ value, with a Pearson coefficient of $R=0.99$. Likewise, nRMSE exhibits a positive correlation with elevation ($R=0.55$). We expect that this correlation can be reduced through more accurate SSI observations in mountainous regions and complex terrain. A lower correlation between nMBE and elevation is observed, with a Pearson coefficient of $R=0.26$.

Finally, Figure \ref{fig:ssi_to_power_residual} illustrates a 2D density histogram of predicted versus measured PV power over the test set, with both values scaled to the [0,1] range for consistency across stations. The test set covers 61 days, yielding roughly 17 million data points. The results indicate strong agreement between predictions and observations, as most points align closely with the 45$^\circ$ line and the linear regression fit shows minimal deviation. A slight overestimation appears at low power values (0–0.2), while at higher power levels (0.5–1) the models tend to underpredict. The symmetry and shape of the marginal histograms confirm a minor low bias and consistent spread between predicted and observed values. Overall, prediction skill remains high, with average nMAE and nRMSE across all stations of 6.2\% and 10.2\%, respectively. 

\begin{figure}[!t]
	\centering
	\includegraphics[width=1\textwidth]{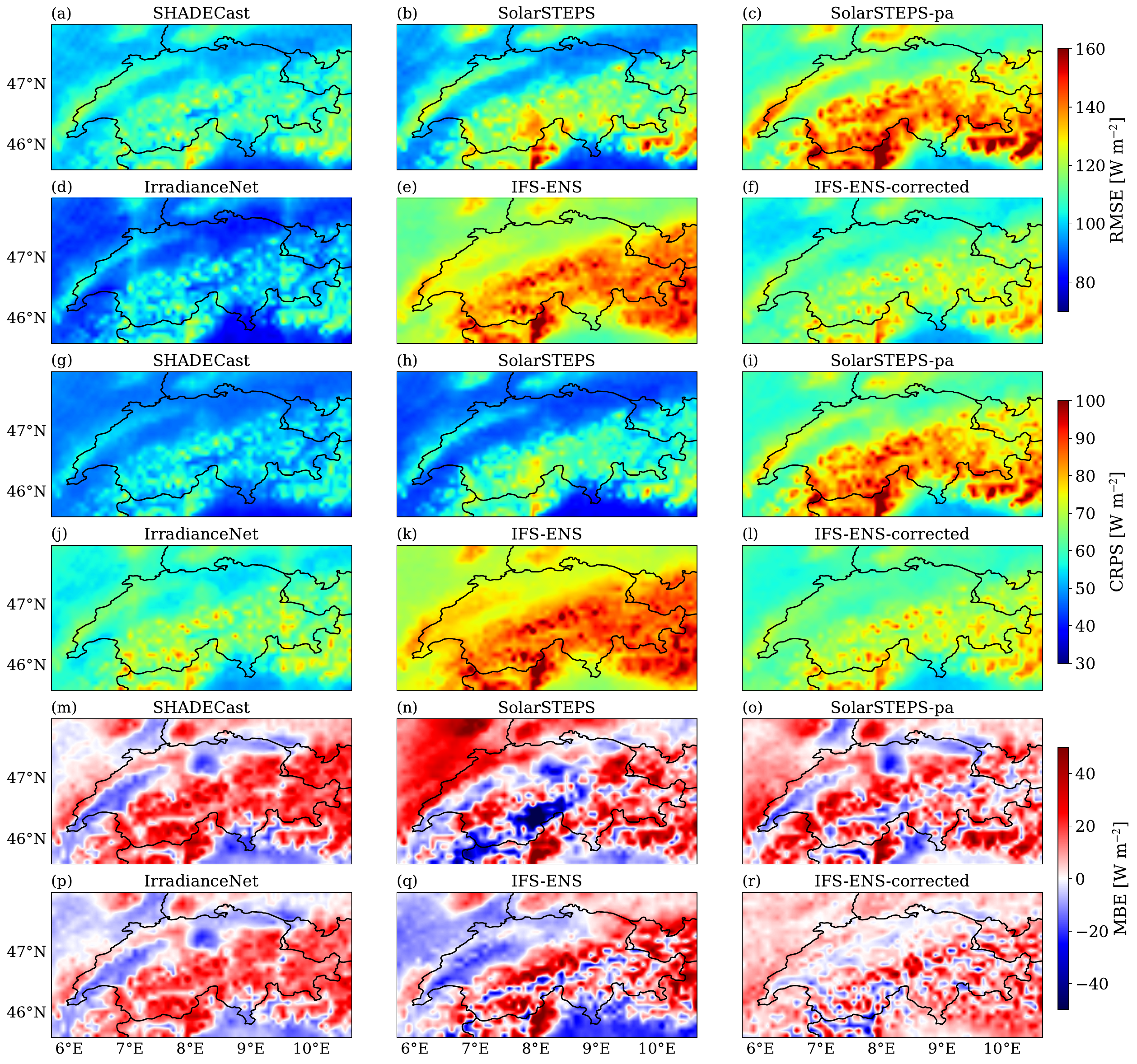}
	\caption{(a-f) RMSE, (g-l) CRPS and (m-r) MBE computed at the pixel level over the study area for all models considered.  Each metric is averaged over all inferences and lead times. The satellite-derived HANNA SSI serves as ground truth. Note that minor artefacts observed near the meridians at approximately 7°E and 9°E in the IrradianceNet fields arise from the patching procedure. Moreover, the CRPS is replaced with the MAE for deterministic models.}
	\label{fig:ssi_comparison_map}
\end{figure}

\subsection{Evaluation of forecast skill for surface solar irradiance}\label{sec:result_ssi}
The satellite-based models evaluated in this study operate at a spatial resolution four times finer than that of IFS-ENS. Therefore, to perform a grid benchmarking (i.e., comparing SSI forecasts at the pixel level over the study area), we first upscale HANNA and the satellite-based SSI inferences by aggregating them from 0.02$^\circ$ to 0.08$^\circ$. Deterministic and probabilistic performance metrics are then computed at each grid point, using HANNA as ground truth, and averaged over all 6158 forecast instances. For probabilistic models, the RMSE, MAE and MBE are computed using the ensemble mean. For deterministic models, CRPS reduces to MAE, and thus the MAE is reported as the corresponding metric. Satellite-based forecasts are generated at the same temporal resolution as HANNA, allowing for a comparison based on instantaneous SSI fields. In contrast, IFS-ENS provides hourly SSI values averaged over the preceding hour. Therefore, comparisons with HANNA are performed using corresponding hourly averages. Further details of the metric definitions are provided in \ref{sec:metrics_app_2}.

The spatial comparison of forecast skill across the area of interest is presented in Figure \ref{fig:ssi_comparison_map}, with results averaged over all lead times and forecast instances. Among all models, IrradianceNet, which is the only deterministic model, achieves the lowest RMSE, corresponding to an average reduction of about 10.5\% compared with the probabilistic ML and optical-flow models and 26.0\% relative to IFS-ENS. However, IrradianceNet shows a comparatively high CRPS (note that CRPS reduces to MAE for deterministic models), while the probabilistic models tend to produce forecasts that are both more reliable and sharper. For instance, the spatially averaged CRPS of SHADECast is 17.7\% lower than that of IrradianceNet. In other words, while the deterministic model excels in point accuracy, the probabilistic models provide more informative and better-calibrated uncertainty estimates.

Within the probabilistic models, SHADECast achieves the lowest CRPS, establishing it as the most reliable and sharp probabilistic SSI forecast model. While SolarSTEPS performs better in low-elevation regions, its skill deteriorates substantially over the Alpine arc, whereas SolarSTEPS-pa, which does not account for cloud evolution, exhibits a substantially weaker performance, with its domain-averaged CRPS being 34.7\% higher than that of SolarSTEPS. All satellite-based models outperform IFS-ENS, which suffers from reduced skill in mountainous terrain.  Notably, IFS-ENS-corrected achieves a 11.5\% RMSE reduction and a 17.3\% decrease in CRPS relative to IFS-ENS, underscoring the effectiveness of bias correction in improving the skill of physics-based probabilistic forecasts.

\begin{figure}[t]
	\centering
	\includegraphics[width=1\textwidth]{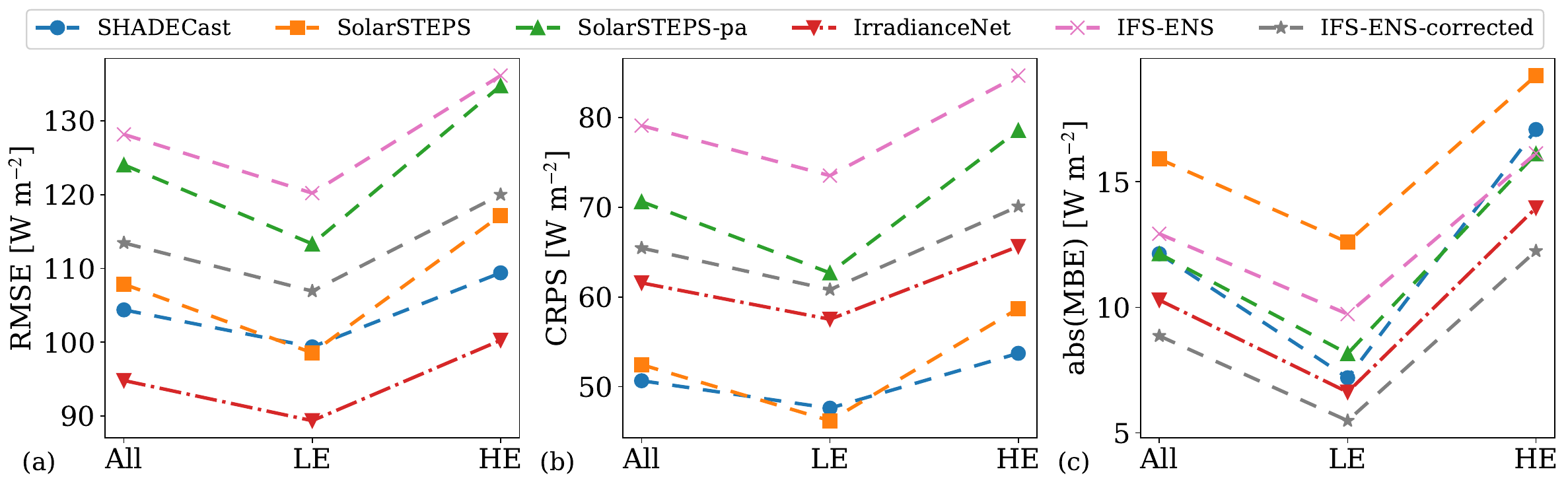}
	\caption{(a) RMSE, (b) CRPS and (c) MBE averaged over all inferences and lead times. Results are additionally averaged over three regions of interest: the full area shown in Figure \ref{fig:ssi_comparison_map} (All), pixels with mean elevation below 790 m (LE), and pixels with mean elevation above 790 m (HE). The elevation threshold was set to the median of the elevation distribution computed over all pixels. Dashed-dotted and dashed lines represent deterministic and probabilistic models, respectively. Moreover, the CRPS is replaced with the MAE for deterministic models.}
	\label{fig:ssi_comparison_elevation}
\end{figure}

Overall, most models exhibit a positive bias, indicating a general tendency to overestimate SSI, with MBE values reaching up to 45 W m$^{-2}$. This overestimation is especially pronounced at higher elevations, while biases at low elevations tend to be negative. IFS-ENS-corrected shows the smallest bias, consistent with the effect of the applied correction.

Figure \ref{fig:ssi_comparison_elevation} provides a global overview of RMSE, CRPS and MBE spatially averaged over three different areas: the entire study area, low-elevation, and high-elevation areas. The distinction between low- and high-elevation regions is based on the median of the elevation distribution measured over all pixels, which is 790 m. We observe that SSI forecasts quality decreases in high-elevation areas. This is expected, as mountainous areas can be characterized by local meteorological processes and lower SSI satellite retrieval quality. Additionally, IFS-ENS operates at a limited spatial resolution, meaning that terrain features such as valleys, ridges, and slopes may be smaller than the model grid. As a result, important sub-grid effects, including terrain shading and slope orientation, are not fully captured, which can lead to reduced forecast accuracy in high-elevation regions. 

\begin{figure}[t]
	\centering
	\includegraphics[width=1\textwidth]{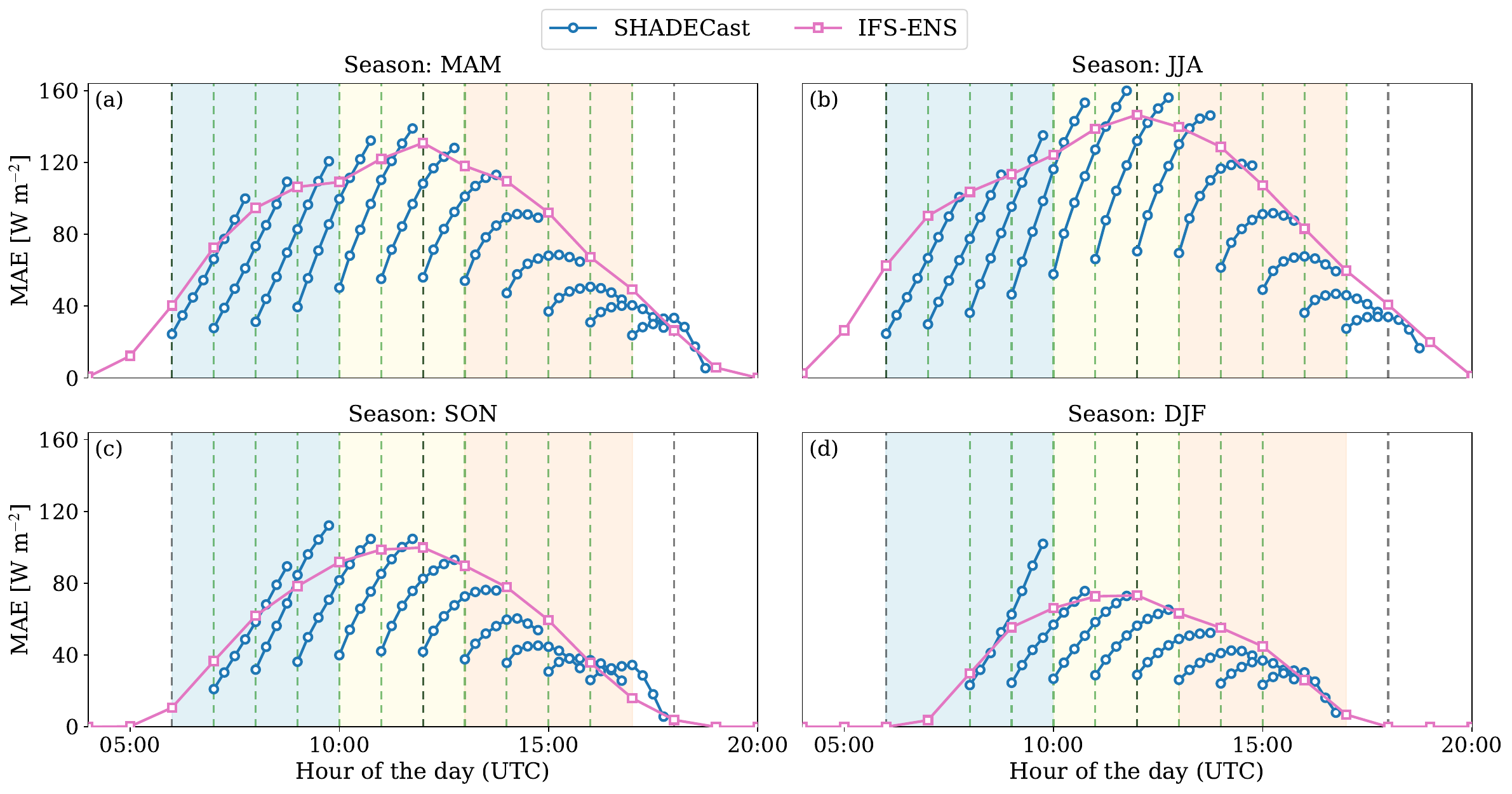}
	\caption{Diurnal evolution of the MAE for SSI forecasts generated with SHADECast (satellite-based) and IFS-ENS (NWP), averaged across all pixels and forecast instances, during the months (a) MAM, (b) JJA, (c) SON and (d) DJF. The vertical dashed green and black lines mark the times at which satellite-based and NWP forecasts are issued, respectively. The light blue, yellow, and orange shaded regions mark the morning, midday, and afternoon periods, respectively.}
	\label{fig:ssi_comparison_line_mae}
\end{figure}

To better understand how forecast skill varies throughout the day, Figure \ref{fig:ssi_comparison_line_mae} illustrates the diurnal patterns in forecast errors for satellite-based and NWP models, averaged over all forecasts and all pixels in the area of interest. The IFS-ENS forecasts exhibit an MAE that closely mirrors the diurnal SSI cycle. Forecast errors are small after sunrise, increase toward midday when irradiance is highest, and then decrease again as sunset approaches. The seasonal dependence is also significant, with MAE values being substantially higher in summer (JJA) than in winter (DJF). Satellite-based forecasts, available at hourly intervals throughout the daylight period, show a different error pattern. At short lead times, SHADECast achieves significantly lower MAE than IFS-ENS because it relies directly on recent satellite observations and thus remains close to ground truth. However, its accuracy degrades as lead time increases, with MAE values often higher than those of IFS-ENS for lead times close to 2 h. In late-afternoon forecasts, the MAE naturally decreases for increasing lead times, as SSI approaches zero near sunset. Similar to the NWP model, the MAE exhibits a parabolic pattern over the course of the day, with the curvature becoming more pronounced as the lead time increases, and with larger errors observed during spring (MAM) and summer (JJA). For clarity, Figure \ref{fig:ssi_comparison_line_mae} reports only results for SHADECast among the satellite-based models. However, all satellite-driven approaches exhibit comparable behavior -- see supplementary material.

\begin{figure}[t]
	\centering
	\includegraphics[width=1\textwidth]{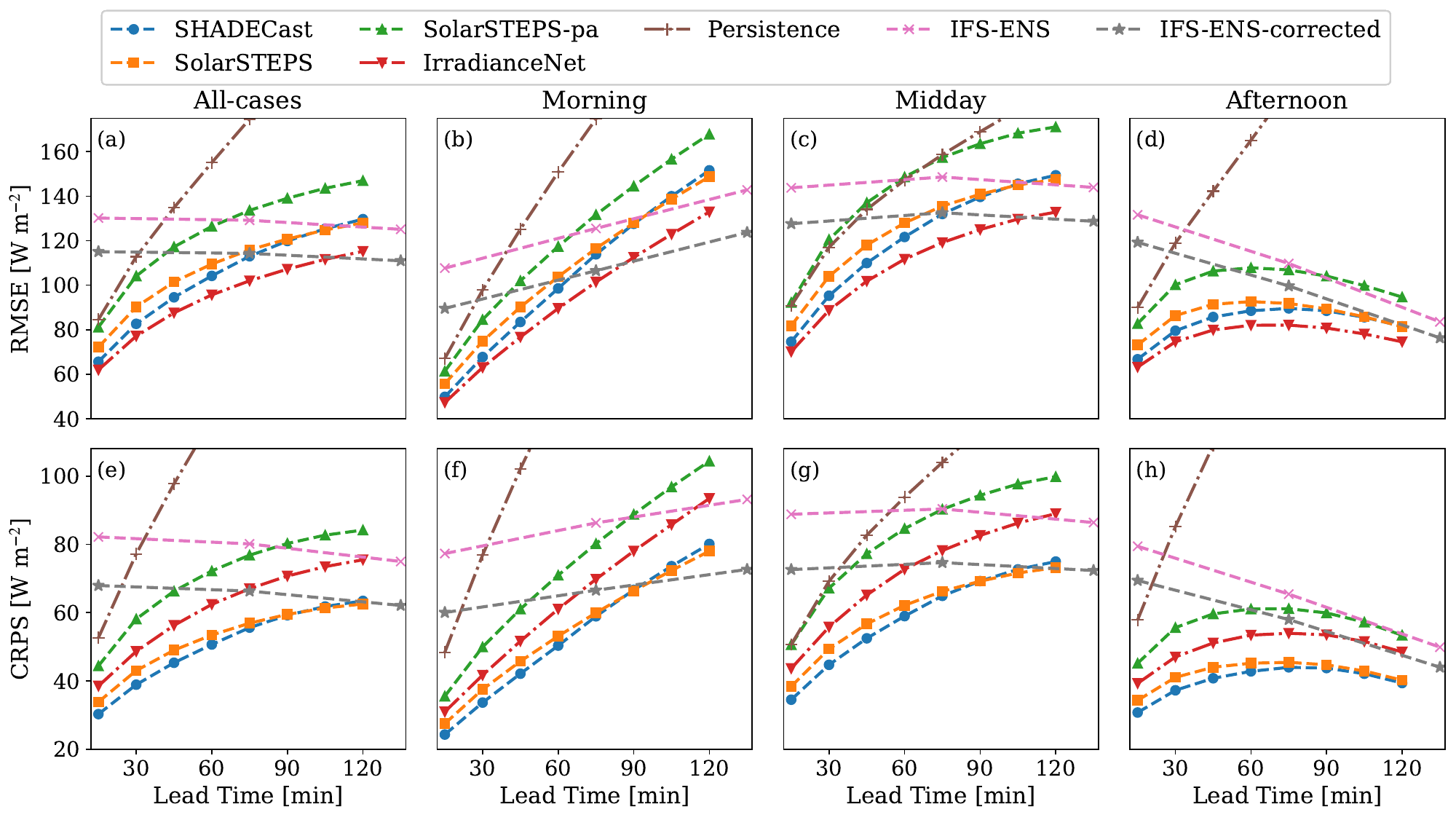}
	\caption{(a-d) RMSE and (e-h) CRPS averaged across the study area and all forecasts for three different periods of the day: morning, midday and afternoon. Results averaged over all daylight hours (All-cases) are also included. The CRPS is replaced with the MAE for deterministic models. Dashed-dotted and dashed lines represent deterministic and probabilistic models, respectively.}
	\label{fig:ssi_comparison_line_crps_time_of_day}
\end{figure}

\begin{figure}[t]
	\centering
	\includegraphics[width=1\textwidth]{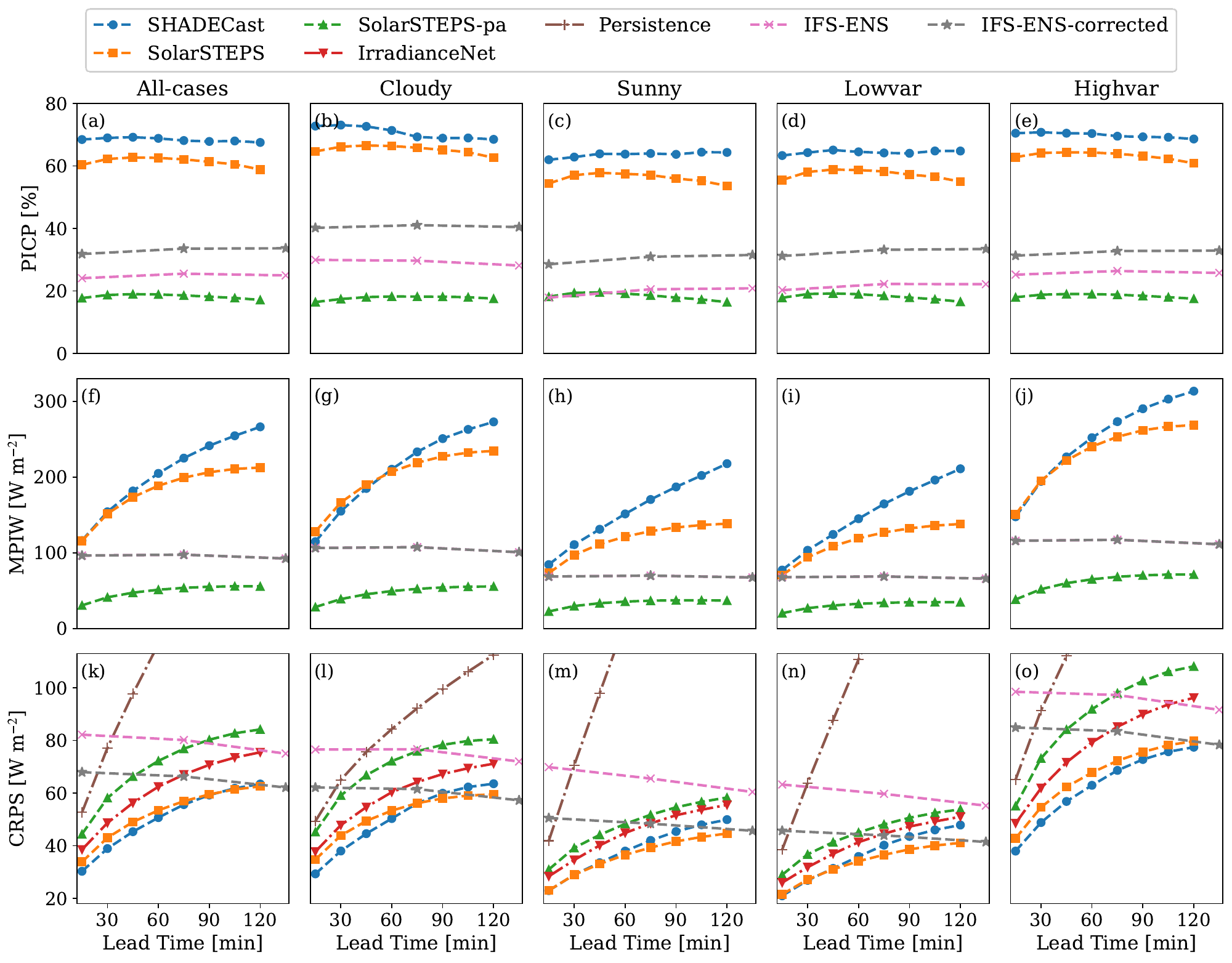}
	\caption{(a-e) PICP, (f-j) MPIW and (k-o) CRPS averaged across the study area and all forecasts for five different weather scenarios: all-cases, cloudy, sunny, low-variability, and high-variability. The PICP and PINAW are only shown for probabilistic models. The CRPS is replaced with the MAE for deterministic models. In panel (k-o), dashed-dotted and dashed lines represent deterministic and probabilistic models, respectively.}
	\label{fig:ssi_comparison_line_crps}
\end{figure}

Figure \ref{fig:ssi_comparison_line_crps_time_of_day} summarizes the evolution of RMSE and CRPS as a function of lead time, grouped into three periods of the day, i.e. morning, midday, and afternoon, as defined in Figure \ref{fig:ssi_comparison_line_mae}. For satellite-based models, both RMSE and CRPS increase steadily during the morning and midday, reflecting the increase in SSI. In the afternoon, these metrics exhibit a parabolic-like behavior, due to the SSI approaching zero near sunset. The largest errors occur at midday, although the rate of error growth is steepest during the morning hours. IFS-ENS displays a different behavior, with RMSE and CRPS more closely following the diurnal cycle of SSI rather than showing a monotonic degradation with lead time. As a result, when averaging over all daylight hours, the resulting error profiles exhibit only a weak dependence on lead time. We note that the bias correction reduces the RMSE of about 11.5\% in the all-cases scenario (i.e. averaged over all available forecasts), demonstrating its effectiveness in improving forecast accuracy. Overall, satellite-based models deliver superior accuracy at short lead times, but their performance degrades more rapidly compared with the NWP model as the forecast horizon extends. 

To assess model performance under different weather regimes, we classify each day into four categories: cloudy, sunny, low-variability (lowvar), and high-variability (highvar). This classification is performed at the daily level, such that all inferences within a given day are assigned to the same category. Specifically, we compute the daily mean and standard deviation of the CSI fields used as input to the satellite-based models. From the resulting distributions across all days, we determine the 25th and 75th percentiles. Days with a mean CSI below the 25th percentile are classified as cloudy, while those above the 75th percentile are classified as sunny. Similarly, days with a standard deviation below the 25th percentile are classified as low-variability, whereas those above the 75th percentile are classified as high-variability. When performance metrics are averaged across all 6158 inferences, we refer to this case as all-cases.

\begin{figure}[!t]
	\centering
	\includegraphics[width=1\textwidth]{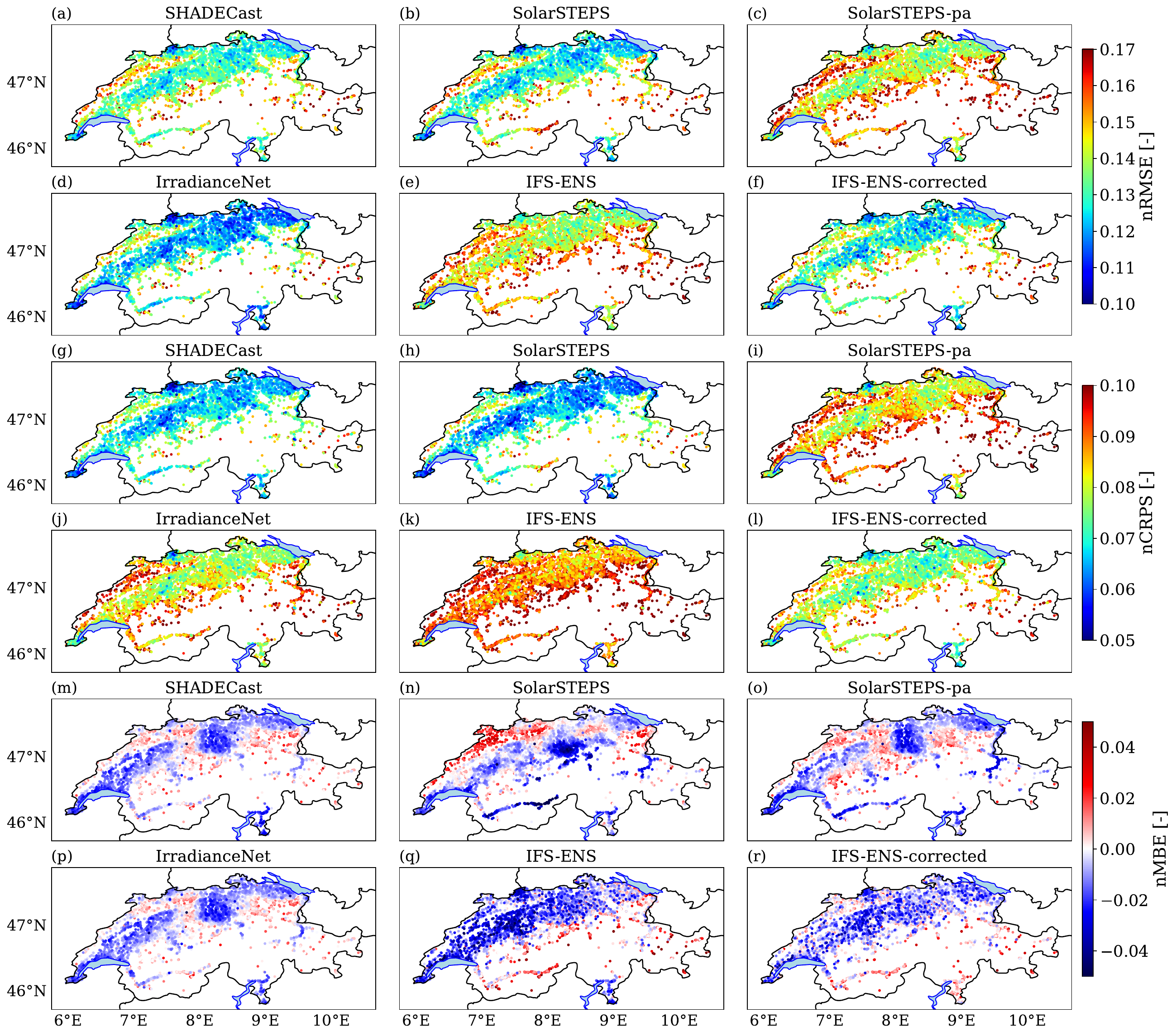}
	\caption{(a-f) nRMSE, (g-l) nCRPS and (m-r) nMBE computed for each PV system with all models considered.  Each metric is averaged over all inferences and lead times. Operational PV production data are used as ground truth. The nCRPS is replaced with the nMAE for deterministic models.}
	\label{fig:power_comparison_map}
\end{figure}

Figure \ref{fig:ssi_comparison_line_crps}(a,f,k) compares the reliability, sharpness, and overall probabilistic performance of all models as a function of lead time averaged over all available forecasts (all-cases). We observe that satellite-based models accounting for cloud evolution are the most reliable but less sharp, with PICP values reaching up to 70\% for SHADECast and 60\% for SolarSTEPS. Moreover, SHADECast exhibits the highest growth rate in the MPIW with increasing lead time, as observed by \cite{Carpentieri2025}. The absence of cloud evolution mechanisms in SolarSTEPS-pa results in an ensemble with limited spread, producing very low PICP and MPIW values at all lead times. These findings are consistent with \cite{Carpentieri2023, Carpentieri2025}. Bias correction applied to the IFS-ENS forecasts increases the PICP by 52\% compared to the non-corrected version, while the PINAW remains unchanged since the ensemble distribution is unaffected by the bias correction -- see \ref{app:bias_correction}. For lead times below 1h, SHADECast achieves a 8.2\% lower CRPS than SolarSTEPS, making it the most accurate model. At lead times near 2 h, SHADECast and SolarSTEPS show similar forecast skill. Although the CRPS of IFS-ENS is higher than that of satellite-based models for small lead times, it shows a lower growth rate. For completeness, the persistence model is also included. This model exhibits a CRPS (MAE) 73.9\% higher than SHADECast at 15 min lead time, and deteriorates sharply at longer lead times.

Figure \ref{fig:ssi_comparison_line_crps} also illustrates results obtained across different weather scenarios. Although the absolute magnitude of the error metrics varies with atmospheric conditions, the relative ranking of the models remains consistent. The PICP exhibits only minor sensitivity to weather type, whereas both the mean prediction interval width and the CRPS vary significantly. For example, under cloudy conditions, the CRPS is about 50\% higher than in sunny conditions. Similarly, high-variability conditions are more challenging than low-variability conditions, therefore showing less accurate forecasts. Notably, SHADECast demonstrates superior skill in high-variability conditions, whereas SolarSTEPS performs better in the low-variability regime. This difference may arise from SolarSTEPS inability to generate accurate forecasts in the presence of time-varying cloudiness distributions, a limitation of its underlying linear AR model \cite{Carpentieri2025}.

\subsection{Evaluation of forecast skill for PV power production}\label{sec:result_power}
First, SSI forecasts are converted into PV power using the procedure described in Section \ref{sec:result_ssi_to_power}. For each forecast, lead time, and ensemble member, SSI values are interpolated to the locations of the PV stations and then used as inputs to the station-specific XGBoost models. Since we have a separate XGBoost model per PV station, this setup amounts to roughly $16 \times 10^9$ inferences, enabled by the computational efficiency of the XGBoost library and the use of multiprocessing. The resulting PV power forecasts are then validated against the operational dataset introduced in Section \ref{sec:pvdata}. Since the PV systems have significantly different nominal capacities, the error metrics are normalized with the station-specific $P_{95}$ value. Further details on the evaluation metrics are provided in \ref{sec:metrics_app_2}.

\begin{figure}[t]
	\centering
	\includegraphics[width=1\textwidth]{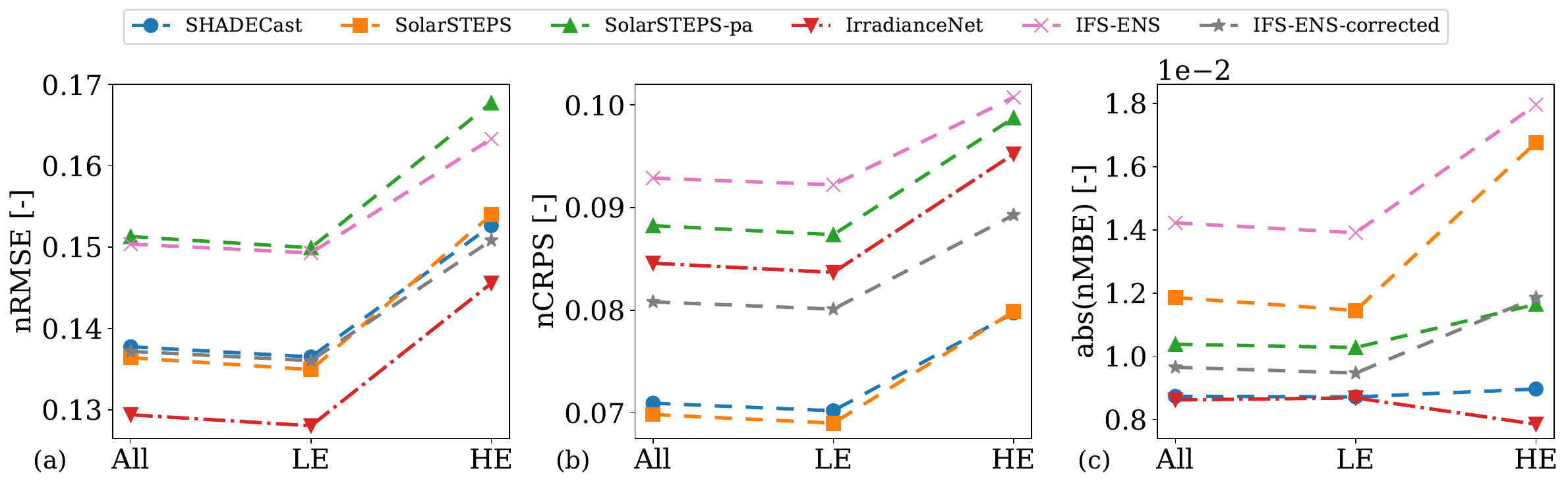}
	\caption{(a) nRMSE, (b) nCRPS and (c) nMBE averaged across all inferences and lead times. Results are further averaged across three groups: all PV systems (All), stations below 790 m elevation (LE), and stations above 790 m elevation (HE). The elevation threshold was set to the median of the elevation distribution computed over all pixels. Dashed-dotted and dashed lines represent deterministic and probabilistic models, respectively. Moreover, the nCRPS is replaced with the nMAE for deterministic models.}
	\label{fig:power_comparison_elevation}
\end{figure}

Figure \ref{fig:power_comparison_map} shows the nRMSE, nCRPS, and nMBE of the power forecasts, evaluated against the operational dataset. Results are averaged over all weather conditions (All-cases scenario) and across all lead times. Since the accuracy of the PV forecast strongly depends on the quality of the underlying SSI forecasts, the spatial patterns of biases and uncertainties are similar to those in Figure \ref{fig:ssi_comparison_map}. In terms of nRMSE, IrradianceNet performs best, with a station-averaged nRMSE 6.1\% lower than that of SHADECast. SHADECast and SolarSTEPS show similar performance overall, although SolarSTEPS achieves lower nRMSE in low-elevation regions. In contrast, the NWP  model performs substantially worse. This shortcoming is also due to its spatial resolution, which is four times coarser than that of satellite-based models, underlining the importance of high-resolution data for site-specific comparisons. 

\begin{figure}[t]
	\centering
	\includegraphics[width=1\textwidth]{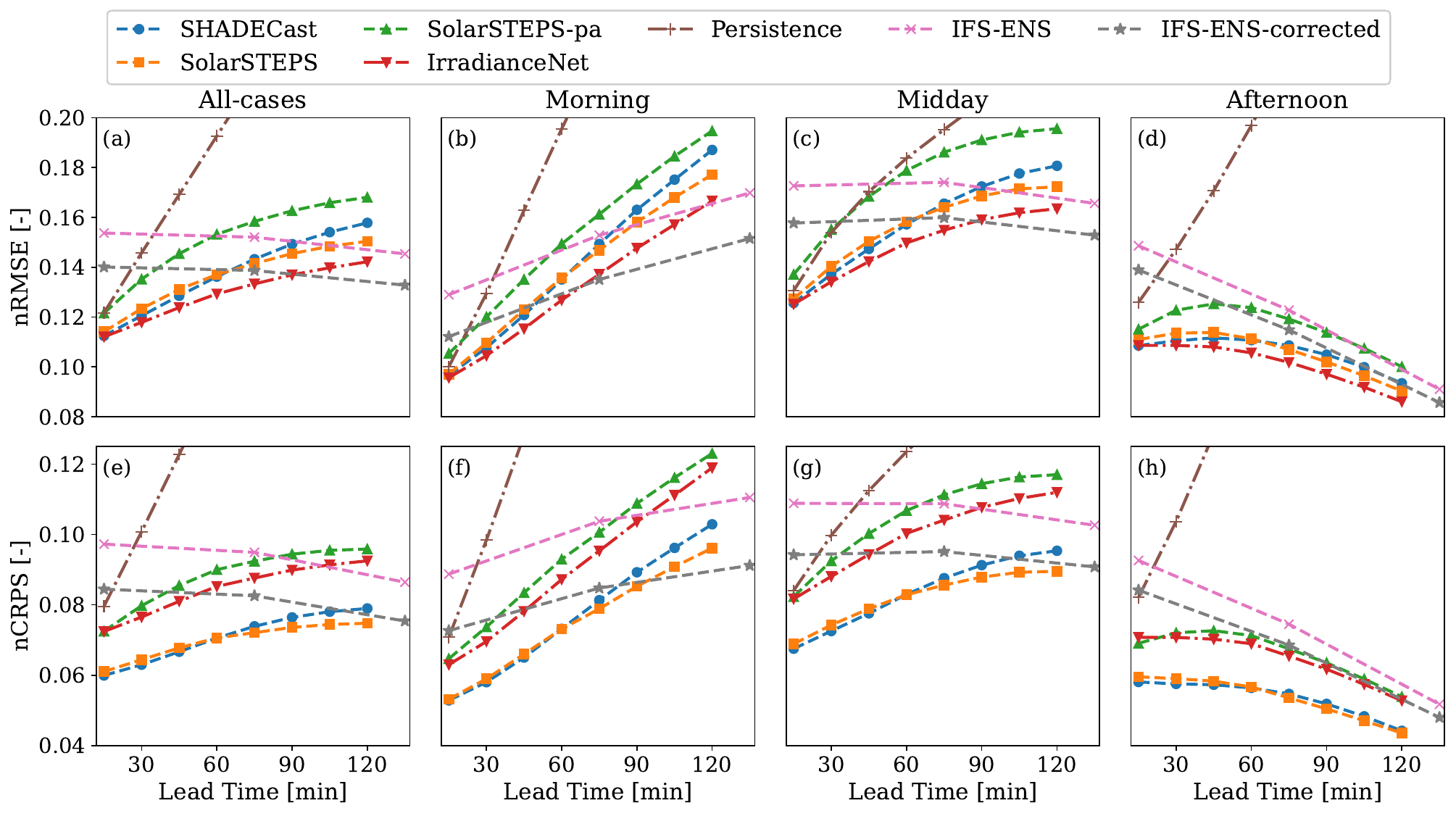}
	\caption{(a-d) nRMSE and (e-h) nCRPS averaged across the study area and all forecasts for three different periods of the day: morning, midday and afternoon. Results averaged over all daylight hours (All-cases) are also included. The CRPS is replaced with the MAE for deterministic models. Dashed-dotted and dashed lines represent deterministic and probabilistic models, respectively.}
	\label{fig:power_comparison_line_crps_time_of_day_p95}
\end{figure}

\begin{figure}[t]
	\centering
	\includegraphics[width=1\textwidth]{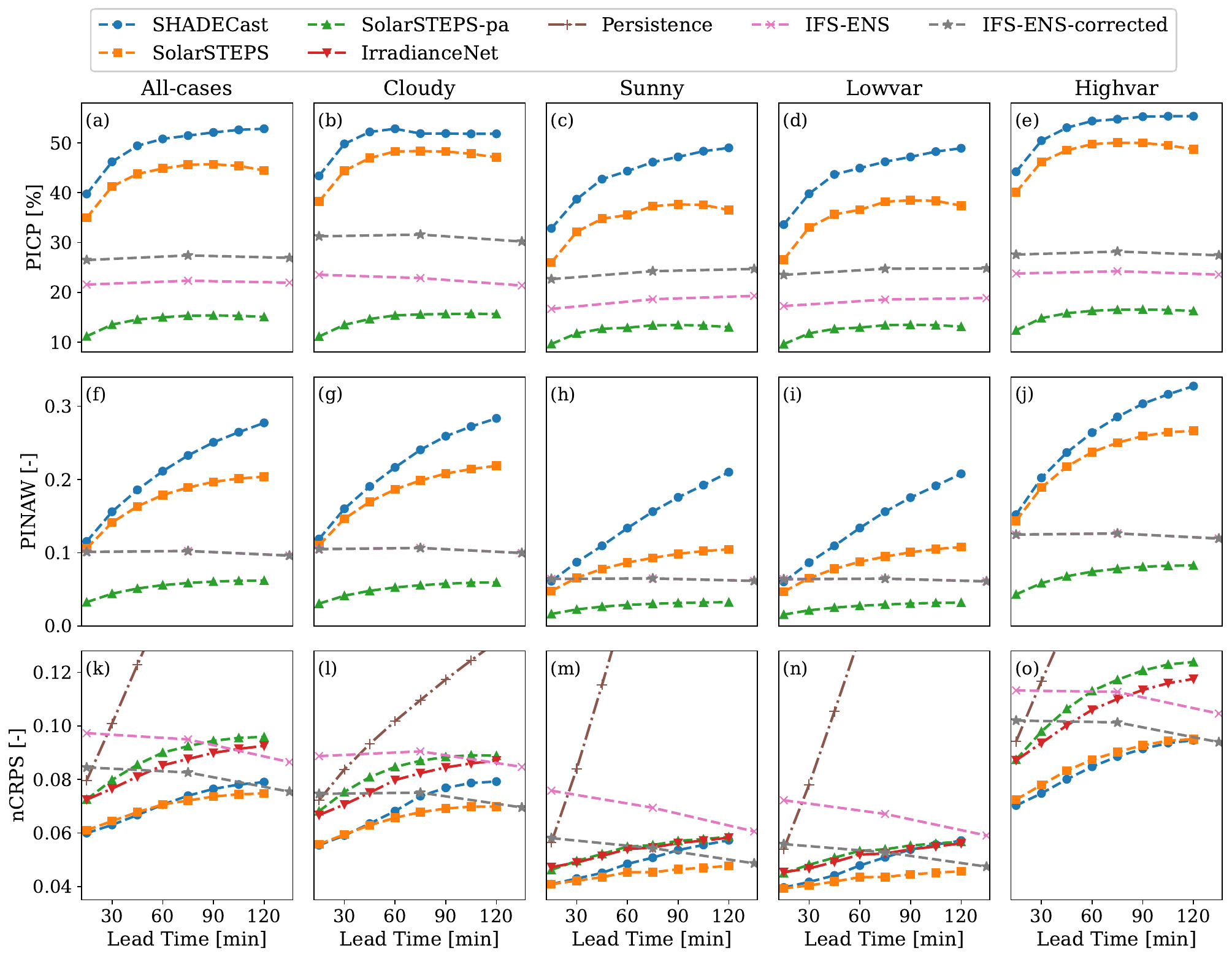}
	\caption{(a-e) PICP, (f-j) PINAW and (k-o) nCRPS averaged across all PV systems and all forecasts for five different weather scenarios: all-cases, cloudy, sunny, low-variability, and high-variability. The PICP and PINAW are only shown for probabilistic models. The nCRPS is replaced with nMAE for deterministic models. In panel (k-o), dashed-dotted and dashed lines represent deterministic and probabilistic models, respectively.}
	\label{fig:power_comparison_line_crps_weathertype_allday_p95}
\end{figure}

A similar spatial pattern is found for the nCRPS, with SHADECast and SolarSTEPS providing the most accurate probabilistic forecasts. The absence of cloud evolution in SolarSTEPS-pa results in a station-averaged nCRPS that is 26.4\% higher than that of SolarSTEPS. IFS-ENS performs worse overall, yielding a station-averaged nCRPS that is 30.9\% higher than that of SHADECast. The pattern observed in the nMBE remains consistent with the one observed for the SSI forecasts, with PV power output underestimated mostly in the low-elevation regions (Figure \ref{fig:ssi_comparison_map}(m-r)).

Overall, Figure \ref{fig:power_comparison_map} shows that PV forecast quality and accuracy decrease in high-elevation regions. To further examine this behaviour, we divided the stations into two groups based on the elevation threshold defined in the previous section (790 m) and computed station-averaged results for each group. Figure \ref{fig:power_comparison_elevation} shows the outcomes for three key metrics. Across all models, nRMSE and nCRPS are 11.8\% and 13.6\% higher in high-elevation regions compared to low-elevation regions, respectively. This decline in performance can be attributed to the lower accuracy of SSI forecasts in high-elevation areas together with the higher variability in the weather conditions, as discussed in Section \ref{sec:result_ssi}. Additionally, the presence of snow can significantly alter PV power generation \cite{Andenaes2018}, an effect currently not captured by forecast models.

\begin{figure}[t]
	\centering
	\includegraphics[width=0.5\textwidth]{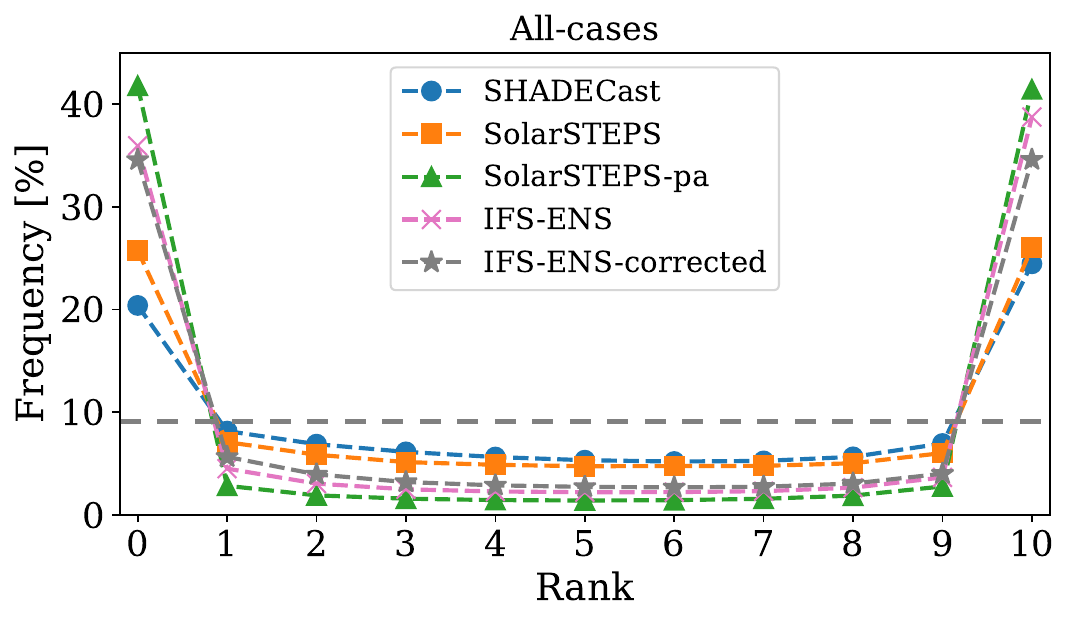}
	\caption{Rank histogram of the probabilistic models, computed over all times in the test set (all-cases scenario) and across all lead times. The horizontal dashed gray line indicates perfect reliability, i.e. uniform distribution.}
	\label{fig:power_comparison_line_rank}
\end{figure}

The PV power forecast errors divided into the three periods of the day are shown in Figure \ref{fig:power_comparison_line_crps_time_of_day_p95}. The generated PV power has a very strong positive correlation with SSI and therefore follows the same diurnal cycle. Hence, the pattern observed is very similar to the one shown in Figure \ref{fig:ssi_comparison_line_crps_time_of_day}. Satellite-based models provide the most accurate forecasts at short lead times, while the IFS-ENS-corrected model achieves the lowest RMSE beyond lead times of 1 h. This behaviour is particularly pronounced during the morning and midday hours, when rapidly changing irradiance conditions cause RMSE and CRPS to increase sharply with lead time for satellite-based models. Next, Figure \ref{fig:power_comparison_line_crps_weathertype_allday_p95} shows the station-averaged reliability, sharpness, and overall probabilistic performance of all models as a function of lead time across the different weather scenarios. In all scenarios, SHADECast remains the most reliable model, but it is also the least sharp, as its ensemble spread grows faster than that of the other models. In terms of nCRPS, SHADECast and SolarSTEPS perform very similarly for lead times below 1 hour, while SolarSTEPS achieves the best performance at longer horizons, particularly in cloudy conditions. The NWP model shows a 50\% higher nCRPS than SHADECast at the 15-minute lead time in the all-cases scenario, although this gap decreases with increasing lead time. While the persistence model achieves a relatively low nCRPS at a 15-minute horizon, its performance degrades rapidly at longer lead times. When evaluating performance across different weather types, a similar pattern as in Figure \ref{fig:ssi_comparison_line_crps} emerges, that is, cloudy and high-variability conditions remain the most challenging for PV power forecasting.

\begin{figure}[t]
	\centering
	\includegraphics[width=1\textwidth]{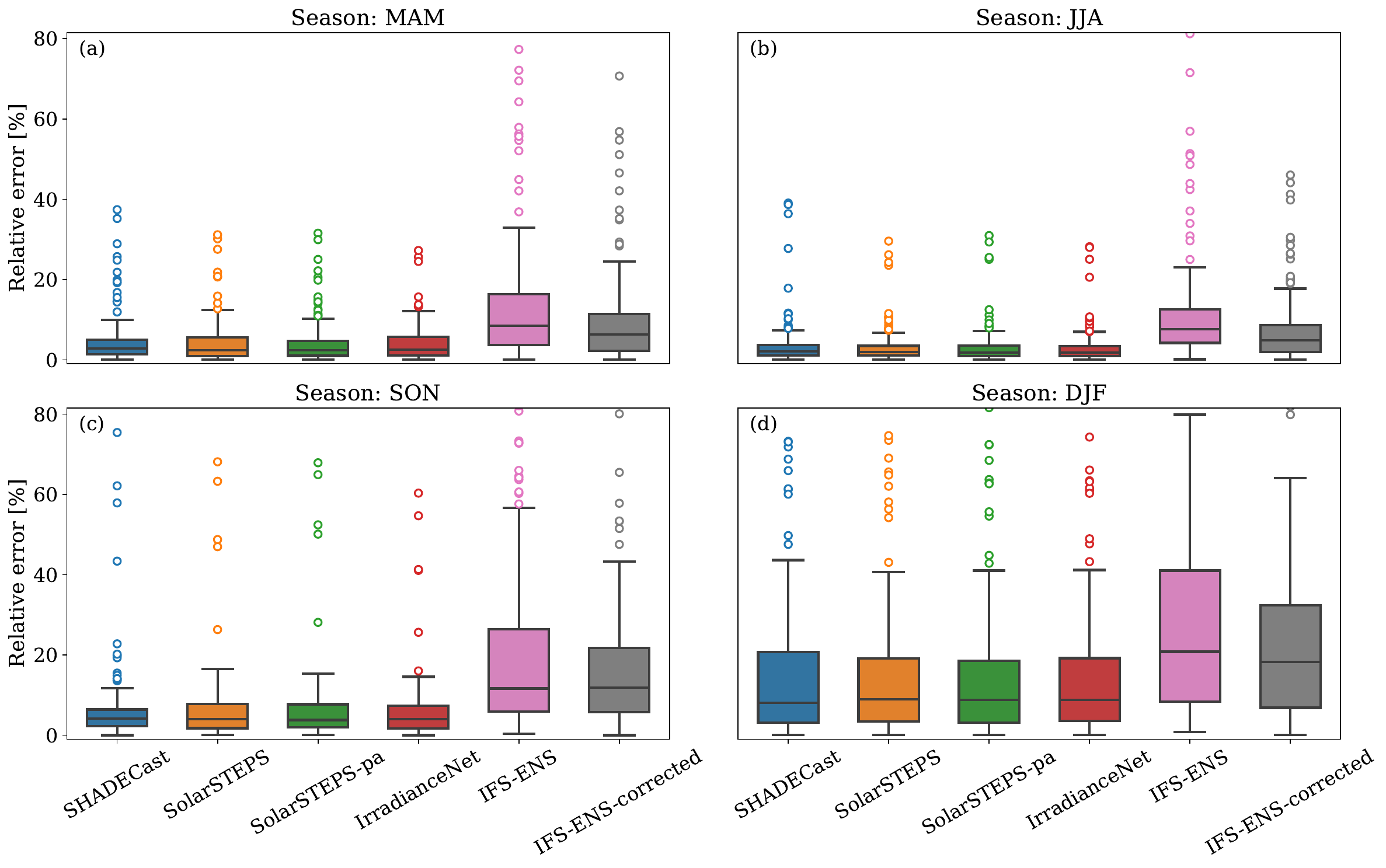}
	\caption{Distribution of the relative error between measured and predicted total (i.e summed over all stations) PV power during the months (a) MAM, (b) JJA, (c) SON and (d) DJF, shown for all models at the 15-minute lead time. To improve readability, the y-axis range excludes a small number of outliers.}
	\label{fig:power_comparison_line_relerr}
\end{figure}

We also evaluated the ensemble forecast performance through rank histograms as shown in Figure \ref{fig:power_comparison_line_rank}. These histograms illustrate how often the observations fall into each rank bin when compared to the sorted ensemble forecasts. For a perfectly reliable ensemble, the distribution should be uniform, indicating that observations are equally likely to fall anywhere within the ensemble spread. However, all models exhibit a U-shaped rank histogram. This indicates that the ensemble spread is too narrow relative to the actual variability of the observations, i.e., the forecasts are underdispersive or overconfident. Consistent with \cite{Carpentieri2025}, SHADECast emerges as the least underdispersive model, followed by SolarSTEPS. By contrast, the NWP model is highly underdispersive, with around 80\% of the measurements lying outside of the ensemble spread.

Finally, we evaluate daily total PV production by summing the predicted power across all installations at the shortest available lead time (15 minutes for satellite-based models) and comparing it with the measured daily output. Applying this procedure to all days in the two-year period yields 726 samples, that is, one relative error value per day. The resulting distributions for each model are shown in Figure \ref{fig:power_comparison_line_relerr}. At this lead time, satellite-based models exhibit similar performance and clearly outperform IFS-ENS. The lowest errors occur in summer (JJA) while winter (DJF) shows the highest errors, likely driven by increased cloudiness, higher variability, and the presence of snow. It is important to note that the reported errors reflect both forecasting inaccuracies and uncertainties in the irradiance-to-power conversion process. Nevertheless, the median relative error for the satellite-based models remains around 2.2\% during the MAM and JJA months, with only minor differences among the models. Moreover, the relative difference between measured and predicted national PV power remains below 1\% for 18\% of the 726 days analyzed, and below 10\% for 82\% of the days for all satellite-based models.

\section{Conclusions}\label{sec:conclusions}
We presented a novel framework for spatiotemporal PV power forecasting and applied it to evaluate the reliability, sharpness, and overall performance of seven PV forecast models. The models rely on a diverse set of SSI forecast approaches: SHADECast and IrradianceNet, two ML-based models, SolarSTEPS, an optical flow model, and IFS-ENS, a physics-based NWP model. While IFS-ENS is a classic NWP model, SHADECast, SolarSTEPS, and IrradianceNet solely rely on satellite observations. For each model, we first generated SSI forecasts for lead times of up to two hours, then used station-specific ML models to convert SSI into PV power, and finally assessed PV forecast skill against measurements from 6434 PV installations across Switzerland. To the best of our knowledge, this is the first spatiotemporal PV forecast framework and the first demonstration of PV forecasting on a large-scale countrywide PV network.

The decision to train a separate XGBoost model for each station was motivated by the partly complex topography of the study area. In addition, differences in panel orientation and inclination across stations introduce site-specific patterns that an ML-based model can effectively capture. Given the strong correlation between SSI and PV production, we found that the accuracy of the irradiance-to-power conversion depends strongly on the quality of the underlying SSI measurements. Consequently, the availability of a dense PV network distributed over a wide area also provides a basis for evaluating the performance of SSI retrieval methods. For example, the HANNA SSI fields are known to have reduced accuracy at higher elevations. Accordingly, we observed a clear correlation between the nRMSE and the stations elevations, with a Pearson coefficient of $R = 0.55$. The irradiance-to-power conversion models achieved an average nMAE of 6.2\% despite the limited set of input features.

The forecast skill of all models was evaluated against HANNA for SSI and against the operational dataset for PV power output. Satellite-based models outperformed the physics-based IFS-ENS model, particularly at short lead times. Two main factors contribute to this behaviour. First, satellite-based models directly rely on recent satellite observations and thus remain close to ground truth, while the NWP model forecast error follows the diurnal cycle of SSI and PV power generation. Second, the satellite-based models operate at a spatial resolution four times finer than IFS-ENS, which proves particularly advantageous when evaluating site-specific performance. Applying a bias correction improved the skill of IFS-ENS forecasts, demonstrating its effectiveness in enhancing forecast accuracy.

We found that SHADECast outperforms SolarSTEPS in SSI forecasting at short lead times and in high-elevation regions, while SolarSTEPS achieves marginally better performance than SHADECast in low-elevation regions. For PV power generation, the two models show comparable performance, although SHADECast produces a more consistent ensemble spread and demonstrates higher reliability. Furthermore, the absence of cloud evolution in SolarSTEPS-pa results in a significant drop in performance compared to SolarSTEPS. These findings emphasize the importance of models that account not only for cloud advection but also for cloud evolution. The deterministic model IrradianceNet consistently achieves the lowest RMSE. However, its MAE is higher than the CRPS of the probabilistic satellite-based models. In other words, the deterministic model excels in point accuracy, but the probabilistic models, despite having somewhat higher RMSE, offer better-calibrated uncertainty estimates. 

Figure \ref{fig:ssi_comparison_line_crps_time_of_day} summarizes the evolution of RMSE and CRPS as a function of lead time, grouped into three periods of the day, i.e. morning, midday, and afternoon, as defined in Figure \ref{fig:ssi_comparison_line_mae}. For satellite-based models, both RMSE and CRPS increase steadily during the morning and midday, reflecting the increase in SSI magnitude. In the afternoon, these metrics exhibit a parabolic-like behavior, due to the SSI approaching zero near sunset. The largest errors occur at midday, although the rate of error growth is steepest during the morning hours. IFS-ENS displays a somewhat different behavior, with RMSE and CRPS following the diurnal cycle of SSI closely, with a somewhat slower deterioration by lead time. As a result, when averaging over all daylight hours, the resulting error profiles exhibit only a weak dependence on lead time. We note that the bias correction reduces the RMSE of about 11.5\% in the all-cases scenario, demonstrating its effectiveness in improving forecast accuracy. Overall, satellite-based models deliver superior accuracy at short lead times, but their performance degrades more rapidly compared with the NWP model as the forecast horizon extends. 

For satellite-based models, forecast errors rise steadily during the morning and midday hours, reflecting the increasing SSI. In the afternoon, errors follow a parabolic pattern as SSI declines toward zero near sunset. The largest errors occur at midday, while the steepest rate of error growth is observed in the morning. In contrast, errors in NWP model forecasts more closely follow the diurnal SSI cycle. Among the four types of weather conditions examined, cloudy and high-variability conditions remain the most challenging for SSI and PV power forecasting, while sunny and low-variability conditions yield substantially better forecast performance. We also found that high-elevation regions exhibit lower forecast skill than low-elevation regions for both SSI and PV power. In addition to terrain effects and Alpine meteorological processes, snow cover can further complicate PV predictions in Alpine regions by modifying surface reflectivity, irradiance, and potentially partially covering PV installations.

At the country level, the total PV power predicted by satellite-based models closely aligns with observations, with relative errors below 10\% for 82\% of the 726 days considered. Relative errors are lowest during summer months, while winter conditions, characterized by higher cloud variability and snow cover, result in larger discrepancies.

In this study, we proposed the first spatiotemporal PV forecast framework and demonstrated it for satellite-based PV power forecasting. Future research should investigate novel data aggregation strategies, for instance at the level of power grid zones. Further, a comparison with regional higher-resolution model forecasts, such as from the Icosahedral Non-hydrostatic (ICON) model operated by MeteoSwiss for the area of this study, would also be of interest, yet forecasts and regional reanalysis were not made openly available. The growing share of solar energy in the power grid increases the need for more accurate forecasts to mitigate operational challenges such as grid instability and reserve power management. To address this, new probabilistic ML-based spatiotemporal PV forecast models can be investigated, with the goal of further reducing short-term grid imbalances and enhancing overall system reliability.

\section*{Declaration of Competing Interest}
The authors declare no conflict of interest.

\section*{Data availability}
The authors do not have permission to share data.

\section*{Acknowledgements}
We acknowledge funding from the Swiss National Science Foundation (grant 200654). We thank Mathieu Schaer and Christian Steger of MeteoSwiss for the support to account for topographic shading using the HORAYZON library. We also thank Anke Tetzlaff and Uwe Pfeifroth of MeteoSwiss and the German Weather Service, respectively, for compiling and providing the HANNA dataset. 

\section*{Supplementary data}
Supplementary material related to this article can be found online. 

\section*{Declaration of generative AI and AI-assisted technologies in the manuscript preparation process}
During the preparation of this work, the authors used ChatGPT in order to help with writing style. After using this tool, the author reviewed and edited the content as needed and takes full responsibility for the content of the published article.

\clearpage

\begin{figure}[t]
	\centering
	\includegraphics[width=1\textwidth]{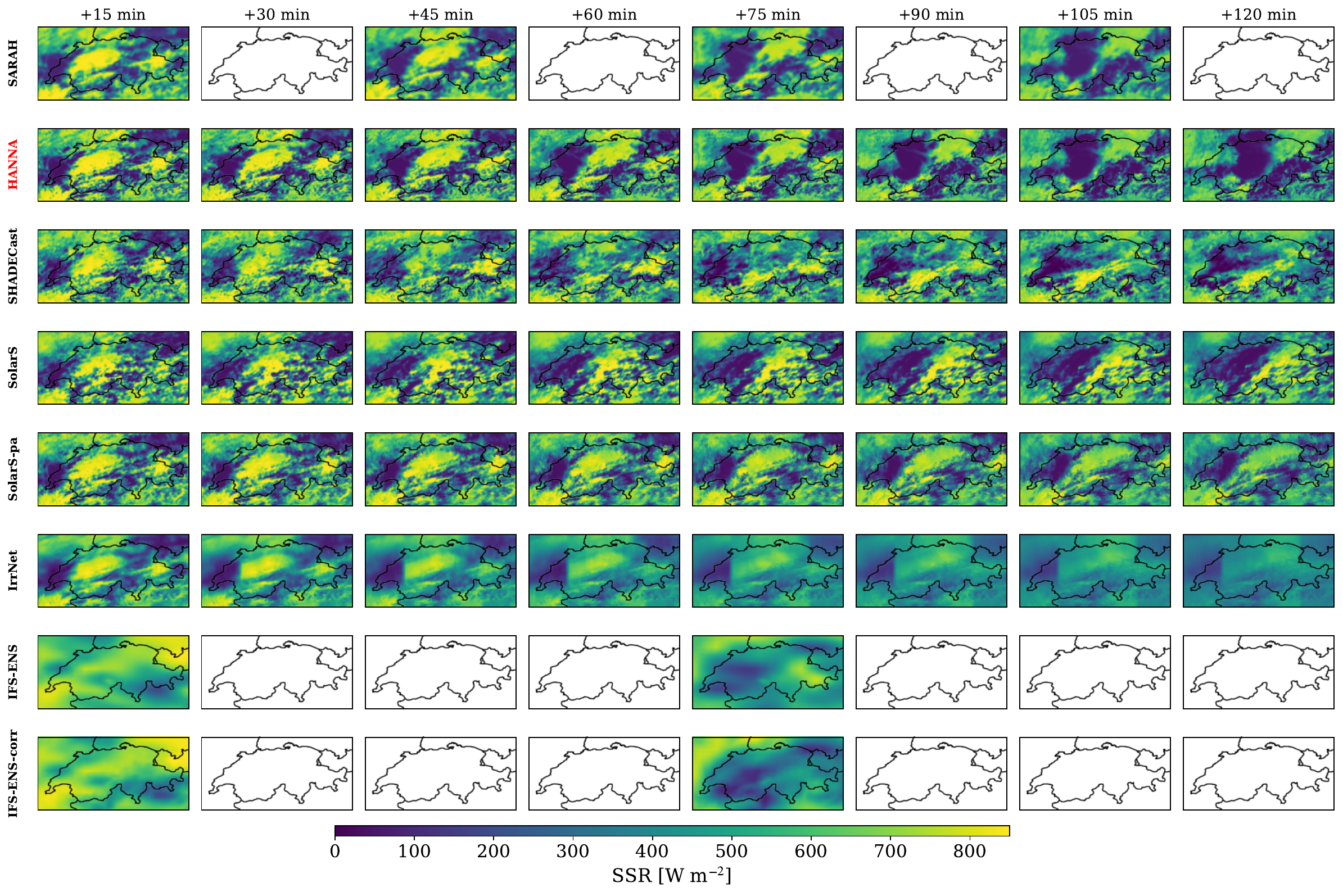}
	\caption{Satellite-based SSI observation and model forecasts over the area of interest at lead times ranging from +15 to +120 minutes. The forecasts are issued at 12:00 UTC on 6 August 2019, a day with high-variability weather. For the probabilistic models, the ensemble member chosen is the one with the lowest RMSE. The black lines denote national borders. The row highlighted with the red label is used as the ground truth. Missing panels reflect differences in the temporal resolution of the satellite observation and forecast models.}
	\label{fig:ssi_forecast_all}
\end{figure}

\appendix
\section{SSI and PV power forecasts}\label{app:ssi_power_forecast}
Figure \ref{fig:ssi_forecast_all} compares the spatial patterns of SSI over the area of interest at lead times ranging from +15 to +120 minutes for the forecast models considered in this study. The forecasts were issued at 12:00 UTC on 6 August 2019. The satellite observations (including both SARAH and HANNA) show pronounced spatial variability, with significant convective cloud development over central Switzerland, localized high-irradiance regions, and sharp gradients associated with cloud structures. Data-driven models capture the observed spatial patterns and their temporal evolution considerably well for lead times up to +60 minutes, with performance declining at longer horizons. In contrast, the NWP-based forecasts (IFS-ENS and IFS-ENS-corrected) produce smoother, more homogeneous SSI fields, reflecting the coarser spatial resolution of the model. Overall, the figure illustrates a gradual loss of spatial detail as the forecast horizon increases, highlighting the advantage of high-resolution and data-driven nowcasting for short-term SSI prediction.

\begin{figure}[t]
	\centering
	\includegraphics[width=1\textwidth]{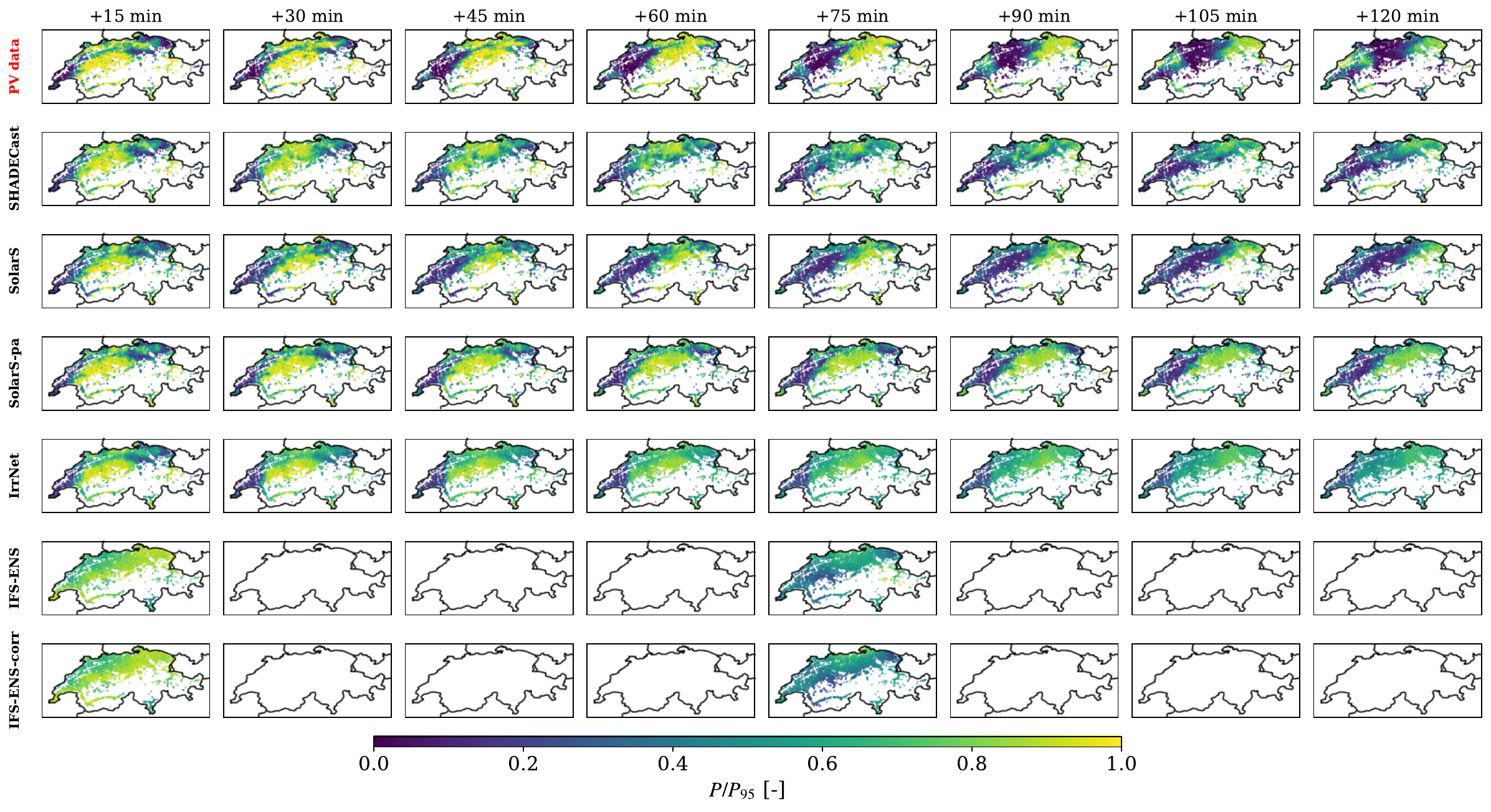}
	\caption{PV power observation and model forecasts over the area of interest at lead times ranging from +15 to +120 minutes. The forecasts are issued at 12:00 UTC on 6 August 2019, a day with high-variability weather. For the probabilistic models, the ensemble member chosen is the one with the lowest RMSE. The black lines denote national borders. The row highlighted with the red label is used as the ground truth. Missing panels reflect differences in the temporal resolution of the satellite observation and forecast models. Note that the PV power output is normalized using the station-specific 95th percentile of the power time series.}
	\label{fig:power_forecast_all}
\end{figure}

Figure \ref{fig:power_forecast_all} shows the corresponding PV power forecasts. The top row presents observed PV power, revealing strong spatial heterogeneity and a well-defined cloud front advected from west to east, resulting in coherent regions of low and high production across Switzerland. All forecast models generally reproduce the observed SSI patterns, with low-irradiance areas corresponding to low PV power generation and vice versa. We note that the selected day is classified as high-variability weather, a scenario in which models show the lowest accuracy. In the supplementary material, we show additional cases.

\section{Surface solar irradiance datasets}\label{app:hanna_sarah_ssi}
The accuracy of SSI observations is crucial for training irradiance-to-power conversion models and for providing ground truth in model benchmarking. In this section, we present a comparison between a widely used open-access dataset, the Surface Radiation Dataset Heliosat (SARAH-3) \cite{Pfeifroth2024}, and a recently released dataset, the High-Resolution European Surface Radiation Data Record (HANNA) \cite{Hanna2025}. This comparison also serves to justify our choice of HANNA as the reference SSI dataset.

The algorithm used for the generation of HANNA follows the HelioMont methodology, which was developed at MeteoSwiss \cite{Hanna2025}. A notable feature of the algorithm is its capacity to differentiate between clouds and snow. This is achieved by leveraging the fact that ground albedo changes slowly over time, whereas cloud albedo varies more rapidly \cite{Stockli2013, Castelli2014}. Additionally, the algorithm also includes cloud shadow corrections \cite{Li2022} and adjustments to account for topographic effects, such as terrain shadowing, surface reflection, local horizon elevation angle, and sky view factor, which are particularly important in mountainous regions \cite{Castelli2014}. SARAH-3 tends to misclassify snow-covered surfaces as clouds, resulting in unreliable SSI estimates in Alpine regions \cite{Carpentieri2023b, Pfeifroth2024}. Additionally, HANNA provides SSI fields at a spatial resolution five times higher than SARAH-3 and at twice the temporal sampling frequency.

\begin{figure}[t]
	\centering
	\includegraphics[width=1\textwidth]{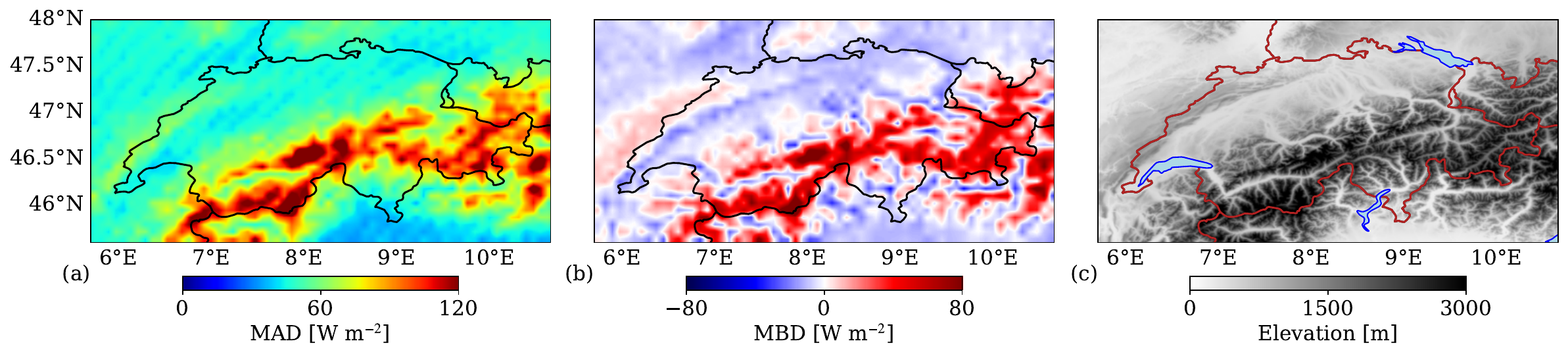}
	\caption{(a) MAD and (b) MBD between SSI estimates from HANNA and SARAH-3 upscaled to the IFS-ENS spatial resolution. Results are averaged over all full-hour timestamps between sunrise and sunset during the two-year period 2019–2020. The MBE is defined as $\mathrm{SSI}_\mathrm{HANNA} - \mathrm{SSI}_\mathrm{SARAH}$. Finally, panel (c) illustrates the elevation map with values in meters above sea level.}
	\label{fig:hanna_sarah_diff}
\end{figure}

Figure \ref{fig:hanna_sarah_diff} compares the mean absolute difference (MAD) and mean bias difference (MBD) between the HANNA and SARAH-3 datasets upscaled to the IFS-ENS spatial resolution. Deviations between HANNA and SARAH-3 remain low in low-elevation regions but increase sharply in mountainous areas such as the Alps and Jura, to the extent that even valleys can be distinguished in the MAD and MBD fields. Part of this deviation is that SARAH-3 tends to misclassify snow as clouds, which leads to persistently underestimated SSI values \cite{Carpentieri2023b}. As a result, the MBD ($\mathrm{SSI}_\mathrm{HANNA} - \mathrm{SSI}_\mathrm{SARAH}$) becomes strongly positive, with differences exceeding 100 W m$^{-2}$ in the Alps.

\section{Metrics}\label{app:metrics}
This appendix provides a detailed description of the metrics used in this study. Specifically, \ref{sec:metrics_app_1} discusses the metrics adopted to evaluate the performance of the irradiance-to-power conversion model, while \ref{sec:metrics_app_2}  presents the metrics used for comparing forecasts in terms of SSI and PV power.

\subsection{Irradiance-to-power conversion}\label{sec:metrics_app_1}
This model takes as input an SSI value and several other geographical and temporal predictors to estimate the PV power output of a specific PV installation at a given time stamp. Here, we denote by $y_{s,n}$ and $\hat{y}_{s,n}$ the measured and predicted PV power at station $s$ and time step $n$, respectively, where $s \in \{1, \dots, S\}$ indexes the PV installations and $n \in \{1, \dots, N\}$ indexes the time steps. The number of stations is fixed to $S=6434$, while $N$ depends on the number of time steps in the considered dataset (training, validation, or test).

For each station $s$, we evaluate three normalized error metrics: the normalized mean absolute error (nMAE$_s$), the normalized root mean square error (nRMSE$_s$), and the normalized mean bias error (nMBE$_s$), defined as
\begin{align*}
\text{nMAE}_s &= \frac{1}{N P_{{95},s}} \sum_{n=1}^{N} \big| \hat{y}_{s,n} - y_{s,n} \big|, \\[2mm]
\text{nRMSE}_s &= \frac{1}{P_{{95},s}} \sqrt{\frac{1}{N} \sum_{n=1}^{N} \big(\hat{y}_{s,n} - y_{s,n}\big)^2}, \\[2mm]
\text{nMBE}_s &= \frac{1}{N P_{{95},s}} \sum_{n=1}^{N} \big(\hat{y}_{s,n} - y_{s,n}\big),
\end{align*}
where $P_{{95},s}$ is the station-specific 95th percentile of the power time series, used here as a normalization factor. A positive value of $\text{nMBE}_s$ indicates systematic overestimation of PV power production by the model, whereas a negative value indicates underestimation. In some applications, the metrics are aggregated across stations by simple averaging:
\begin{equation*}
M = \frac{1}{S} \sum_{s=1}^{S} M_s,
\end{equation*}
where $M$ denotes a generic metric (nMAE, nRMSE, or nMBE in this case). For simplicity, we use the same notation to refer both to the station-wise metric $M_s$ and to its average across stations $M$. We note that these definitions apply to Section \ref{sec:result_ssi_to_power}.

\subsection{SSI and PV power comparison}\label{sec:metrics_app_2}
This section illustrates the scores adopted to compare forecasts in terms of SSI and PV power. To start, we consider forecasts of PV power for a generic PV installation, which consists of $L=8$ lead times and $E=10$ ensemble members. We denote with $\hat{y}_{s,n,e,l}$ the predicted PV power at station $s$, time step $n$, ensemble member $e$ and lead time $l$. Here, $s \in \{1, \dots, S\}$ indexes the PV installations, $n \in \{1, \dots, N\}$ indexes the time steps, $e \in \{1, \dots, E\}$ indexes the ensemble member and $l \in \{1, \dots, L\}$ indexes the lead time. The corresponding observed PV power at station $s$ and lead time $l$, associated with forecast $n$, is denoted by $y_{s,n,l}$.

Deterministic performance metrics are evaluated with respect to the ensemble mean, defined as
\begin{equation*}
\overline{\hat{y}}_{s,n,l} = \frac{1}{E} \sum_{e=1}^{E} \hat{y}_{s,n,e,l}.
\end{equation*}
Deterministic models have $E=1$, therefore $\overline{\hat{y}} = \hat{y}$.

The normalized station-wise metrics averaged over all lead times and all forecasts, shown for instance in Figure \ref{fig:power_comparison_map}, are defined as
\begin{align*}
\text{nMAE}_s &= \frac{1}{N L f_{s}} \sum_{n=1}^{N} \sum_{l=1}^{L} \big| \overline{\hat{y}}_{s,n,l} - y_{s,n,l} \big|, \\[2mm]
\text{nRMSE}_s &= \frac{1}{f_{s}} \sqrt{ \frac{1}{N L} \sum_{n=1}^{N} \sum_{l=1}^{L} \big( \overline{\hat{y}}_{s,n,l} - y_{s,n,l} \big)^2 }, \\[2mm]
\text{nMBE}_s &= \frac{1}{N L f_{s}} \sum_{n=1}^{N} \sum_{l=1}^{L} \big( \overline{\hat{y}}_{s,n,l} - y_{s,n,l} \big),
\end{align*}
where $f_{s} = P_{95,s}$. We note that it is important to apply this normalization, otherwise stations with higher $P_{95,s}$ values would inherently exhibit higher errors, as illustrated in Figure \ref{fig:ssi_to_power_scatter}(a). 

In addition to deterministic scores, we also evaluate metrics tailored to probabilistic forecasts. For each forecast, the ensemble $\{\hat{y}_{s,n,e,l}\}_{e=1}^E$ can be used to construct predictive intervals or to approximate the full predictive distribution.  
We consider three standard metrics: the prediction interval coverage probability (PICP), the prediction interval normalized average width (PINAW), and the normalized continuous ranked probability score (nCRPS).

We define the central $(1-\alpha)$ prediction interval at station $s$, forecast $n$, and lead time $l$ via the empirical quantiles of the ensemble
$\{\hat{y}_{s,n,e,l}\}_{e=1}^E$:
\begin{equation*}
\big[\,\hat{y}^{\mathcal{L}}_{s,n,l},\; \hat{y}^{\mathcal{U}}_{s,n,l}\,\big]
=\big[\,Q_{\alpha/2}(\{\hat{y}_{s,n,e,l}\}_{e=1}^E),\; Q_{1-\alpha/2}(\{\hat{y}_{s,n,e,l}\}_{e=1}^E)\,\big],
\end{equation*}
where $Q_{p}$ denotes the empirical $p$-quantile. The quantity $\alpha$ denotes the nominal risk level, i.e. the probability mass outside the central prediction interval. In this work, we set $\alpha=0.1$, which corresponds to a nominal $90\%$ prediction interval. The coverage probability for station $s$ is then defined as
\begin{equation*}
\text{PICP}_s = \frac{1}{N L} \sum_{n=1}^{N} \sum_{l=1}^{L} \mathds{1}\!\left( \hat{y}^{\mathcal{L}}_{s,n,l} \leq y_{s,n,l} \leq \hat{y}^{\mathcal{U}}_{s,n,l} \right),
\end{equation*}
where $\mathds{1}(\cdot)$ denotes the indicator function.
  
The average width of the same prediction intervals, normalized by the scaling factor $f_{s}$, is given by
\begin{equation*}
\text{PINAW}_s = \frac{1}{N L f_s} \sum_{n=1}^{N} \sum_{l=1}^{L} \left( \hat{y}^{\mathcal{U}}_{s,n,l} - \hat{y}^{\mathcal{L}}_{s,n,l} \right).
\end{equation*}

The nCRPS compares the predictive cumulative distribution function $F_{s,n,l}$, estimated from the ensemble, against the observation $y_{s,n,l}$, and is defined as
\begin{equation*}
\text{nCRPS}_s = \frac{1}{N L f_s} \sum_{n=1}^{N} \sum_{l=1}^{L} 
\int_{-\infty}^{\infty} \big( F_{s,n,l}(z) - \mathds{1}(y_{s,n,l} \leq z) \big)^2 \, dz.
\end{equation*}
In practice, $F_{s,n,l}$ is approximated from the empirical distribution of the ensemble members $\{\hat{y}_{s,n,e,l}\}_{e=1}^E$.

Both deterministic and probabilistic metrics can be further averaged across all stations, yielding the global nMAE, nRMSE, nMBE, PICP, PINAW, and nCRPS, as illustrated in Figure~\ref{fig:power_comparison_elevation}. Depending on the application, the same metrics may instead be averaged across all forecasts and stations for a fixed lead time, as shown in Figure~\ref{fig:power_comparison_line_crps_weathertype_allday_p95}. For clarity, we explicitly state in the text which averaging procedure is applied, while retaining a consistent notation throughout the manuscript. These metrics apply to Section~\ref{sec:result_power}.

When the comparison is carried out in terms of SSI, the same evaluation framework applies. The only difference is that $\hat{y}_{p,n,e,l}$ and $y_{p,n,l}$ denote the predicted and observed SSI values, and the station index $s$ is replaced by the pixel index $p \in \{1, \dots, P\}$, where $P$ is the total number of pixels covering the area of interest. In this case, we set $f_s = 1$, ensuring that the resulting metrics remain expressed in W m$^{-2}$. These metrics are adopted in Section~\ref{sec:result_ssi}.

\section{ML-based bias correction methodology for IFS-ENS}\label{app:bias_correction}
To investigate whether bias correction could improve performance, we applied an ML–based correction to the IFS-ENS SSI forecasts. This is to mitigate systematic errors in the model outputs by learning the relationship between forecast and observed SSI values \cite{Dongjin2020, Lei2021}. This section provides a detailed description of the adopted methodology and model architecture.

The bias correction model was trained using historical forecasts from IFS-ENS and corresponding SSI observations from HANNA for the period 2019–2020. The training, validation, and test sets were defined following the procedure outlined in Section \ref{sec:ssitopower}. Specifically, the two-year period was divided into consecutive 12-day blocks. For each block, ten days were assigned to the training set, while the remaining 2 days were allocated to the validation and test sets.

The bias correction was implemented using a U-Net–based convolutional neural network. The model was configured with nine input channels containing the ensemble-mean forecast, elevation, latitude, longitude, year, and the sine and cosine transformations of both the DoY and HoD. Since temporal features such as DoY and HoD are scalar quantities, they were broadcast to the spatial domain and represented as two-dimensional grids in order to be compatible with the convolutional architecture. The architecture and hyperparameters were optimized with Optuna. The selected configuration consisted of a U-Net with two encoder–decoder levels, with 64 and 128 convolutional filters in the first and second layer, respectively, and incorporating dropout regularization with a rate of 0.19 to mitigate overfitting. Batch normalization and max-pooling operations were used to improve training stability and feature representation. The network was trained with an initial learning rate of $1.5 \times 10^{-3}$ and an adaptive learning rate schedule, which reduced the rate by a factor of 0.12 after six consecutive epochs without improvement on the validation set. The model was optimized using the mean squared error loss function.

The trained model estimates the bias of the ensemble-mean IFS-ENS forecast. This correction is subsequently applied to all ensemble members, thereby adjusting the ensemble mean while preserving the ensemble spread. As a result, the bias-corrected IFS-ENS retains the same PINAW but exhibits a different PICP compared to the original forecasts.

\clearpage

\bibliographystyle{elsarticle-num} 
\bibliography{BibTex}

\end{document}